%% file: main.tex
\documentclass{defaltmlr}
\input{math_commands.tex}

\usepackage{microtype}       
\usepackage{graphicx}        
\usepackage{booktabs}        
\usepackage{wrapfig}         
\usepackage{enumitem}        
\usepackage{multirow}        
\usepackage{makecell}        
\usepackage{ulem}            
\usepackage{pifont}          
\usepackage{bbding}          
\usepackage[linesnumbered,ruled,vlined]{algorithm2e}

\usepackage{url}
\usepackage[table]{xcolor}
\usepackage[dvipsnames]{xcolor}
\usepackage{graphicx}
\usepackage{geometry}
\usepackage{booktabs}
\usepackage[T1]{fontenc}
\usepackage{fontawesome}
\usepackage{fancyhdr}
\usepackage{tocloft}
\usepackage{enumitem}
\usepackage{etoc}
\usepackage{titletoc}
\usepackage{tcolorbox}
\usepackage{listings}
\newcommand\DoToC{%
  \startcontents
  \printcontents{}{1}{\noindent \textbf{\large{Table of Contents}}\vskip3pt\vskip5pt}
  \vskip3pt\vskip5pt
}

\usepackage{amsmath}
\usepackage{amssymb}
\usepackage{mathtools}       
\usepackage{amsthm}          
 \usepackage{textcomp}
 
\usepackage{xcolor}          
\usepackage[most]{tcolorbox} 

\newcommand{\ie}{\textit{i}.\textit{e}.}
\newcommand{\eg}{\textit{e}.\textit{g}.}

\def\model{\texttt{ST-Prune}~}

\definecolor{BlueGreen}{RGB}{105,204,207}
\definecolor{RedOrange}{RGB}{255,69,0}
\newcommand{\blue}[1]{$_{\color{BlueGreen}\downarrow #1}$}
\newcommand{\red}[1]{$_{\color{RedOrange}\uparrow #1}$}
\definecolor{myred}{RGB}{184,26,15}
\definecolor{mydarkred}{rgb}{0.6,0,0}
\definecolor{myblue}{HTML}{268BD2}

\definecolor{bestcolor}{HTML}{2263aa}
\definecolor{secondcolor}{HTML}{2da02d}

\newcommand{\firstres}[1]{{\textcolor{bestcolor}{\textbf{#1}}}}
\newcommand{\secondres}[1]{{\textcolor{secondcolor}{\underline{#1}}}}

\def\PemsEight{\textit{PEMS08}}
\def\UrbanEV{\textit{UrbanEV}}
\def\LargeST{\textit{LargeST}}

\def\HardRandom{\textit{Hard Random}}
\def\CD{\textit{CD}}
\def\Herding{\textit{Herding}}
\def\Kmeans{\textit{K-Means}}
\def\LeastConfidence{\textit{Least Confidence}}
\def\Entropy{\textit{Entropy}}
\def\Margin{\textit{Margin}}
\def\Forgetting{\textit{Forgetting}}
\def\GraNd{\textit{GraNd}}
\def\Cal{\textit{Cal}}
\def\Glister{\textit{Glister}}
\def\GraphCut{\textit{GraphCut}}
\def\FaLo{\textit{FaLo}}
\def\SoftRandom{\textit{Soft Random}}
\def\ThetaGreedy{\textit{$\epsilon$-greedy}}
\def\UCB{\textit{UCB}}
\def\InfoBatch{\textit{InfoBatch}}

\def\GWNet{\textit{GWNet}}
\def\STAEformer{\textit{STAEformer}}
\def\STID{\textit{STID}}
\def\Opencity{\textit{OpenCity}}

\usepackage{wrapfig}

\hypersetup{linkcolor=cyan}
\usepackage{mathtools}
\usepackage{ulem}
\usepackage{pifont}
\usepackage{bbding}

\usepackage{tcolorbox}
\usepackage{makecell}

\makeatletter
\let\orig@fnsymbol\@fnsymbol
\def\@fnsymbol#1{\ifcase#1\or\relax\else\orig@fnsymbol{#1}\fi}
\makeatother

\title{Learning from Complexity: \\Exploring Dynamic Sample Pruning of Spatio-Temporal Training}
\vspace{4cm}

\author{
\parbox{\textwidth}{
Wei Chen\textsuperscript{1,2}, Junle Chen\textsuperscript{2}, Yuqian Wu\textsuperscript{1}, Yuxuan Liang\textsuperscript{1$^*$}, Xiaofang Zhou\textsuperscript{2$^*$}
}
}

\affiliation{\textsuperscript{1}HKUST(GZ), \textsuperscript{2}HKUST}

\abstract{
Spatio-temporal forecasting is fundamental to intelligent systems in transportation, climate science, and urban planning. However, training deep learning models on the massive, often redundant, datasets from these domains presents a significant computational bottleneck. Existing solutions typically focus on optimizing model architectures or optimizers, while overlooking the inherent inefficiency of the training data itself. This conventional approach of iterating over the entire static dataset each epoch wastes considerable resources on easy-to-learn or repetitive samples. In this paper, we explore a \textit{novel training-efficiency techniques, namely learning from complexity with dynamic sample pruning, \model, for spatio-temporal forecasting}. Through dynamic sample pruning, we aim to intelligently identify the most informative samples based on the model's real-time learning state, thereby accelerating convergence and improving training efficiency. Extensive experiments conducted on real-world spatio-temporal datasets show that \model significantly accelerates the training speed while maintaining or even improving the model performance, and it also has scalability and universality.
}

\correspondence{onedeanxxx@gmail.com,$^*$yuxliang@outlook.com,$^*$zxf@cse.ust.hk}
\date{\sffamily Jan 28, 2026}

\begin{document}

\maketitle

\makeatletter
\let\@fnsymbol\orig@fnsymbol
\makeatother

\input{sections/01_introduction}

\input{sections/02_relatedwork}

\input{sections/03_preliminary}

\input{sections/04_methodology}

\input{sections/05_experiments}
\input{sections/06_conclusion}

\clearpage

\definecolor{textgray}{HTML}{6E6E73}
\makeatletter
\newcommand\applefootnote[1]{%
  \begingroup
  \renewcommand\thefootnote{}%
  \renewcommand\@makefntext[1]{\noindent##1}%
  \footnote{#1}%
  \addtocounter{footnote}{-1}%
  \endgroup
}
\makeatother

\bibliography{ref}
\bibliographystyle{plainnat}
\newpage
\appendix

\input{appendix/0_appendix_list}

\end{document}

%% file: math_commands.tex
\usepackage{amsmath,amsfonts,bm}

\def\eqref#1{equation~\ref{#1}}

\def\1{\bm{1}}

\DeclareMathAlphabet{\mathsfit}{\encodingdefault}{\sfdefault}{m}{sl}
\SetMathAlphabet{\mathsfit}{bold}{\encodingdefault}{\sfdefault}{bx}{n}



%% file: sections/01_introduction.tex
\section{Introduction}

With the rapid advancement of sensing technologies, massive volumes of spatio-temporal data are being collected and leveraged in data-driven forecasting scenarios, enabling critical services ranging from traffic management to weather forecasting and power grid operations~\citep{avila2020data,nguyen2023climax}. To capture the complex nonlinear dynamics inherent in such data, spatio-temporal neural networks~\citep{jin2023spatio,jin2024survey} have become the dominant and most powerful paradigm. However, while research community has largely focused on designing increasingly sophisticated architectures for marginal performance gains~\citep{shao2023exploring}, \textit{a fundamental and costly bottleneck has been overlooked: the training process itself.}

Standard spatio-temporal training protocols require iterating over the entire samples in every training epoch~\citep{liu2024largest}. As a result, the widely adopted benchmarks~\citep{li2018diffusion,song2020spatial} are typically restricted to limited spatial regions and temporal spans, severely limiting the scalability of spatio-temporal nueral networks training and inflating their computational cost. This raises a natural research question: 
\textit{do we really need to compute over all available spatio-temporal samples during training stage?}

To answer this question, we first conduct a detailed analysis of the widely used spatio-temporal benchmark (\textsc{PeMS08}). As shown in Figure~\ref{fig:motivation}, the left panel visualizes the Pearson correlation matrix across spatial nodes, revealing that the vast majority exhibit high similarity ($\geq0.8$). Even node pairs with lower similarity still display recurring periodic temporal patterns, as illustrated in the middle panel. Principal component analysis (PCA) along both spatial and temporal dimensions (right panel) further shows that a small number of components are sufficient to reconstruct most of the variance. \textit{Together, these statistics reveal a high degree of spatio-temporal redundancy, which opens up new opportunities for more efficient utilization of training data.}

\input{tables/motivation}

Recent data-centric acceleration research~\citep{zha2025data} have devoted significant effort to identifying unbiased or representative subsets within training datasets. Typical techniques include data pruning~\citep{raju2021accelerating, wang2023brave, qin2023infobatch, zhang2024graph, zhang2024gder, wang2025winning}, dataset distillation~\citep{nguyen2021dataset, wang2022cafe, cazenavette2022dataset, zhao2023dataset, li2023attend, zhang2024navigating}, and coreset selection~\citep{har2004coresets, chen2009coresets, toneva2018empirical, shim2021core}, which respectively retain, synthesize, or select a smaller yet information-rich subset from the original data. Despite their promise, these methods are primarily developed for computer vision and natural language processing tasks and \textit{fail to fully exploit the unique redundancy characteristics of spatio-temporal data highlighted above. }

\begin{figure*}[t!]
    \centering
    \includegraphics[width=1.0\linewidth]{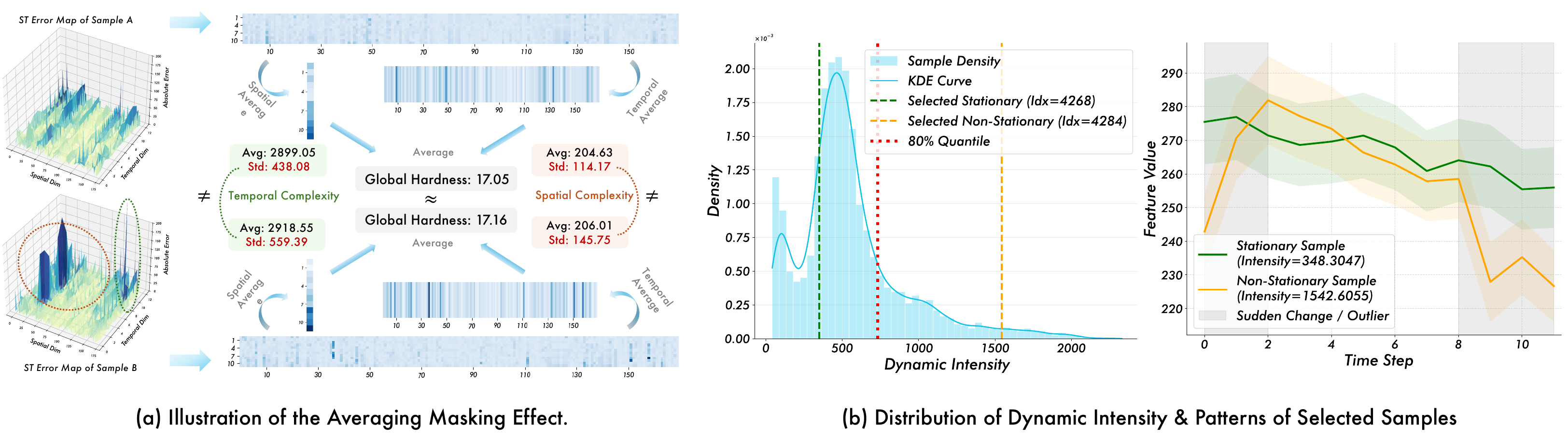}
    \caption{Further analysis of spatio-temporal data insights. (a) Averaging Masking Effect: Low-error nodes dilute critical localized anomalies, necessitating a structural scoring mechanism beyond simple mean error. (b) Long-tail Stationarity Distribution: The dominance of stationary patterns motivates our stationarity-aware rescaling to prevent distribution shift and maintain representativeness.}
    \label{fig:insight}
\end{figure*}

To address this critical gap, we propose \model, a novel dynamic sample pruning framework specifically tailored for spatio-temporal training. Distinct from traditional paradigms that passively process the entire dataset, \model actively curates high-value data subsets to optimize training efficiency. Our fundamental insight is that directly applying generic dynamic pruning strategies~\citep{raju2021accelerating, qin2023infobatch,moser2025coreset} to spatio-temporal data proves ineffective. This failure stems from two specific phenomena: the \textit{Averaging Masking Effect} (Figure~\ref{fig:insight} a) , where critical localized failures are obscured by low global errors, and the inherent \textit{Long-tail Stationarity Distribution} (Figure~\ref{fig:insight} b), where standard pruning induces severe distribution shifts. To this end, \model operationalizes these insights via two innovative components: complexity-informed scoring metric, which incorporates a spatio-temporal heterogeneity penalty to identify structural samples that appear ''globally trivial yet locally intractable``; and stationarity-aware gradient rescaling, which adaptively adjusts weights based on dynamic intensity. This ensures that the preserved samples maintain an unbiased representation of the original data distribution while discarding substantial redundant stationary samples. Ultimately, this design not only significantly mitigates computational overhead but also ensures the robust capture of diverse and complex spatio-temporal patterns. In summary, our contributions are:

\begin{itemize}[leftmargin=*,itemsep=0em]
    \item We propose \model, a dynamic sample pruning method of spatio-temporal training, that shifts the spatio-temporal research focus from solely optimizing the model to intelligently optimizing the data flow during training.
    \item We design a novel framework consisting of a complexity-informed difficulty metric to assess sample informativeness in real-time and an stationarity-aware distribution rescaling to ensure stable and effective training.
    \item Extensive experiments on multiple real-world spatio-temooral datasets demonstrate that \model drastically reduces training time while maintaining or even improving the predictive accuracy of various backbones, proving its effectiveness, efficiency, and universality.
\end{itemize}

%% file: tables/motivation.tex
\begin{figure*}[t!]
\centering
\begin{minipage}[t]{0.3\linewidth}
  \centering
  \includegraphics[width=\linewidth]{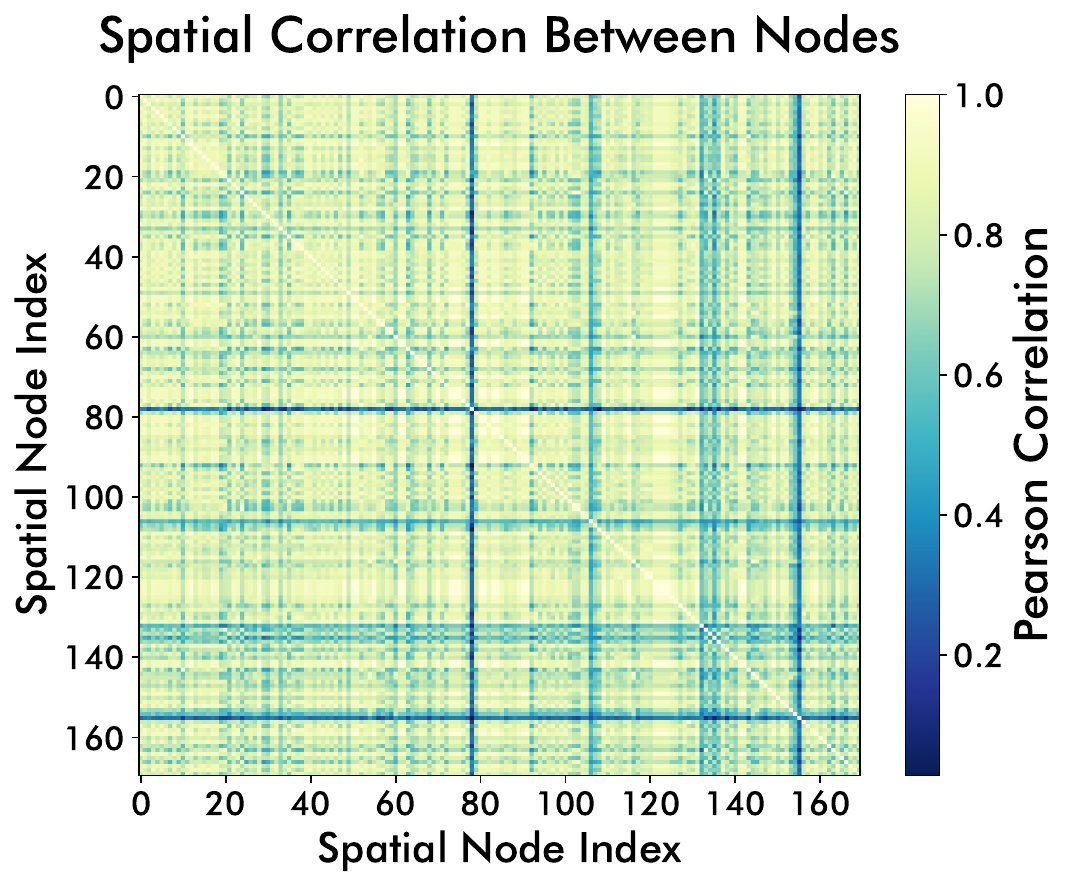}
\end{minipage}\hfill
\begin{minipage}[t]{0.3\linewidth}
  \centering
  \includegraphics[width=\linewidth]{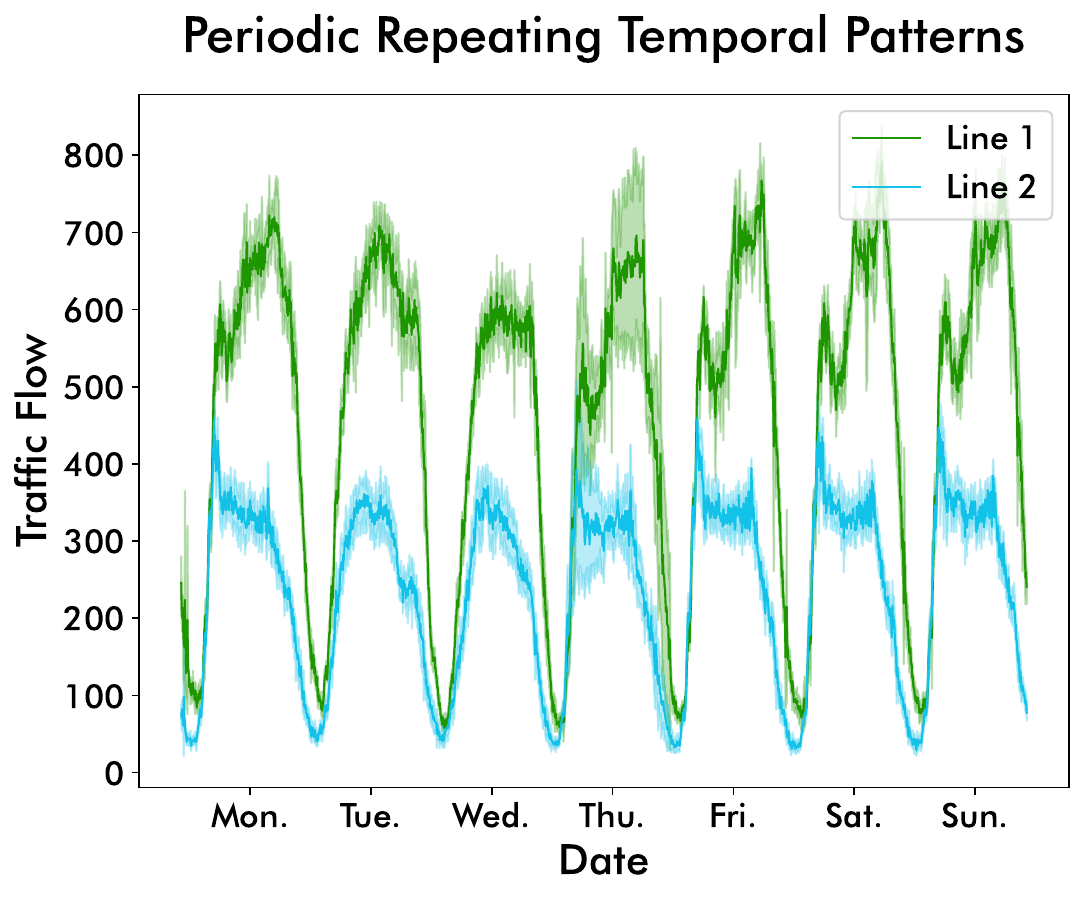}
\end{minipage}\hfill
\begin{minipage}[t]{0.3\linewidth}
  \centering
  \includegraphics[width=\linewidth]{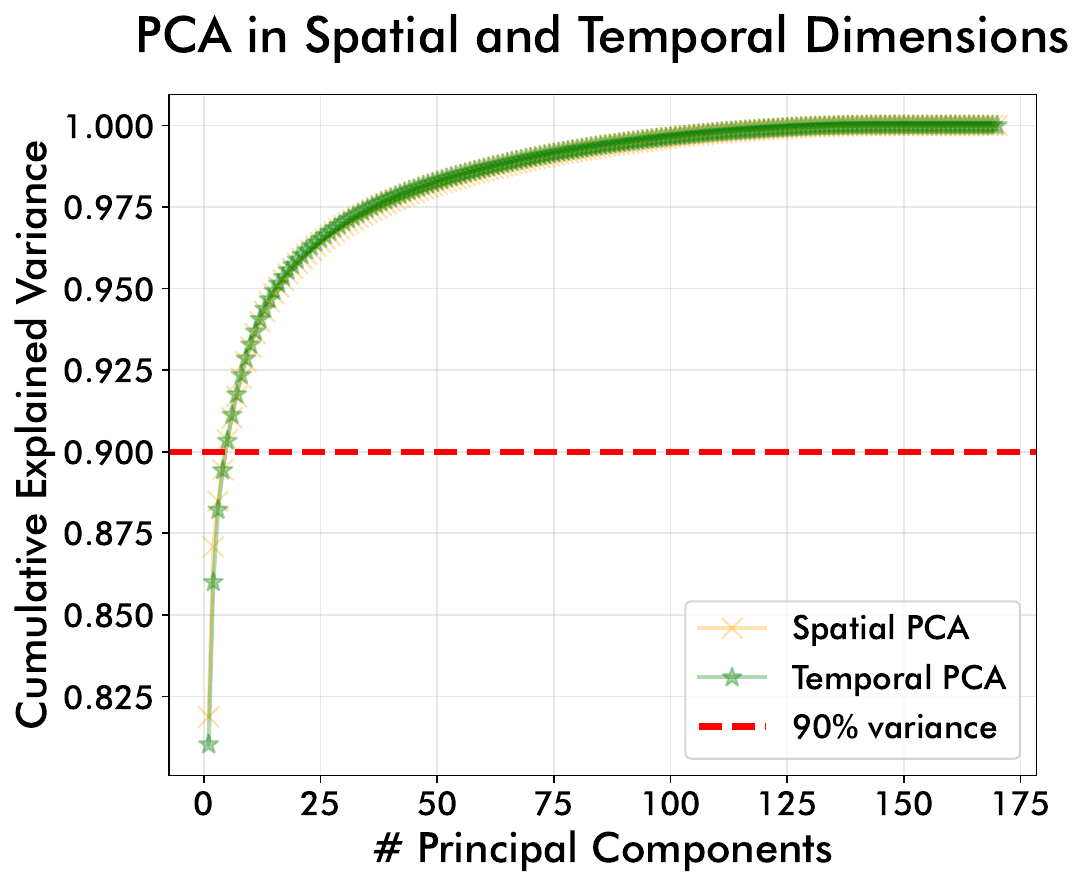}
\end{minipage}
\caption{The spatio-temporal data redundancy characteristics and statistical properties along the spatial and temporal dimensions, exemplified by the \textsc{PeMS08}~\citep{song2020spatial} dataset. For more statistical information on other datasets, please refer to Appendix \ref{appendix_datasets}.}
\label{fig:motivation}
\end{figure*}

%% file: sections/02_relatedwork.tex
\section{Related Work}
\textbf{Efficient Spatio-Temporal Training.}
Spatio-temporal forecasting tasks typically rely on neural networks that integrate spatial and temporal operators~\citep{chen2025expand,chen2025learning} to capture spatial dependencies and temporal dynamics. However, the large number of nodes and extended time horizons~\citep{liu2024largest,yin2025xxltraffic} result in substantial computational costs during training. Previous studies have sought efficiency improvements through model-level optimizations, such as prior structure~\cite{shao2022spatial,cini2023taming,fu2025integration}, graph sparsification~\citep{wu2025dynst,ma2025less}, and model distillation~\citep{tang2024easyst, chen2025information}, or sample-level transformations, including subgraph~\citep{jiang2023spatio,liu2024reinventing,wang2024make,weng2025let,zhao2026fast} and input window~\citep{fang2025efficient,ockerman2025pgt,shao2025blast} sampling. While these approaches reduce computational demand at the model or optimization level, they rarely provide explicit control over the number or distribution of training samples and often tightly couple network component design. \textit{In contrast, we aim to develop a general dynamic spatio-temporal sample pruning strategy that accelerates the training of arbitrary spatio-temporal neural networks while maintaining competitive performance.}

\textbf{Data-Centric Acceleration Strategy.}
Data-centric acceleration strategies~\citep{zha2025data} have drawn considerable attention because they enhance model training and inference efficiency from a data management perspective. Representative approaches include data distillation~\citep{lei2023comprehensive,yu2023dataset}, data pruning, and data selection~\citep{moser2025coreset}. Data distillation seeks to compress the original data distribution into a smaller yet information-equivalent synthetic dataset to reduce training cost, whereas data pruning and selection reduce computational burden by removing redundant samples or selecting the most representative subsets. However, these methods are primarily designed for vision~\citep{wang2018dataset} or language~\citep{albalak2024survey} tasks and struggle to exploit the high redundancy and strong correlations inherent in spatio-temporal data. Moreover, they are often static and sensitive to sample size~\citep{guo2022deepcore}, which may limit generalization under dynamically evolving spatio-temporal distributions. \textit{Notably, our method exploits the complexity of ST samples and dynamically adjusts sample selection to maximize computational efficiency.}

%% file: sections/03_preliminary.tex
\section{Preliminaries}

\noindent \textbf{Probelm Definition (Spatio-Temporal Forecasting).}
From an optimization perspective, spatio-temporal forecasting aims to solve an empirical risk minimization problem. Given a large-scale spatio-temporal training dataset $\mathcal{D} = \{(X_i, Y_i)\}_{i=1}^{|\mathcal{D}|}$ (and an optional graph structure prior $\mathcal{G}$), the objective is to find the optimal model parameters $\theta^*$ by minimizing a global loss function $\mathcal{J}(\theta; \mathcal{D})$:
\begin{equation}\label{eq1}
    \min_{\theta} \mathcal{J}(\theta; \mathcal{D}) = \frac{1}{|\mathcal{D}|} \sum_{i=1}^{|\mathcal{D}|} \mathcal{L}(f_\theta(X_i; \mathcal{G}), Y_i),
\end{equation}
where $f_\theta$ is the forecasting spatio-temporal neural network, and $\mathcal{L}$ is a loss metric (\eg, MAE). $(X_i \in \mathbb{R}^{N \times T_p \times F}, Y_i \in \mathbb{R}^{N \times T_f \times F})$ represent the sensor readings of $C$ measurement features at $N$ spatial locations over $T_p$ past and $T_f$ future consecutive time steps, respectively.
This formulation treats every sample $(X_i, Y_i)$ equally and implies that the gradient for each training epoch is computed over the entire, often redundant, dataset $\mathcal{D}$. This exhaustive summation is the primary source of computational inefficiency in the spatio-temporal training process.

\noindent \textbf{Task Definition (Dynamic Sample Pruning).}
We formulate dynamic sample pruning as an optimization problem aimed at approximate the convergence of the objective in Equation~\ref{eq1}. Instead of using the full dataset $\mathcal{D}$ at each epoch $e$, we seek to identify an optimal training subset $\mathcal{D}_e^* \subset \mathcal{D}$ with a constrained size $|\mathcal{D}_e| = k_e < |\mathcal{D}|$. The model parameters are then updated based only on this subset: $\theta_e = \operatorname{Update}(\theta_{e-1}, \mathcal{D}_e)$. Ideally, the optimal subset $\mathcal{D}_e^*$ should be the one that provides the steepest descent direction for the true, underlying data distribution $\mathcal{P}$:
\begin{equation}
    \mathcal{D}_e^* = \operatorname*{argmin}_{\mathcal{D}_e \subset \mathcal{D}, |\mathcal{D}_e|=k_e} \mathbb{E}_{(X, Y) \sim \mathcal{P}}[\mathcal{L}(f_{\theta_e}(X; \mathcal{G}), Y)].
\end{equation}
However, minimizing this objective is intractable as the true distribution $\mathcal{P}$ is unknown. However, we have empirically shown above that this distribution exhibits a significant low-rank property, that is, the training samples are highly redundant. Thus, the core challenge of dynamic sample pruning is to design a tractable and efficient proxy strategy to select an informative subset $\mathcal{D}_e$ at each epoch.

%% file: sections/04_methodology.tex
\section{Methodology}

\begin{figure*}[htbp!]
    \centering
    \includegraphics[width=1.0\linewidth]{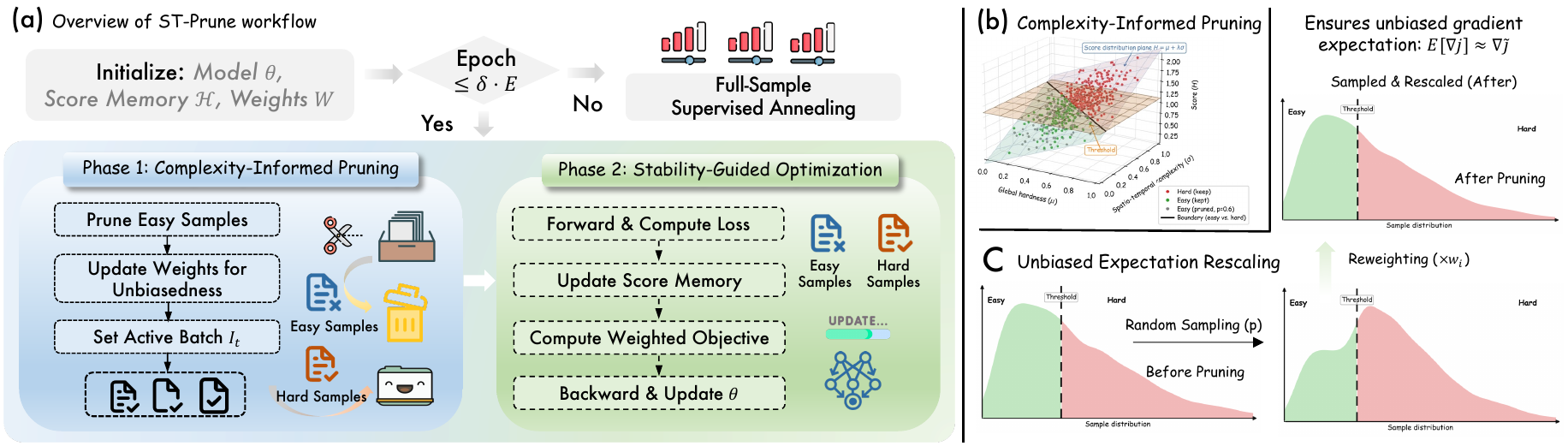}
    \caption{The overall workflow of the \model for efficient data pruning during spatio-temporal training.}
    \label{fig:overall}
\end{figure*}

\subsection{Complexity-Informed Pruning}

\subsubsection{Beyond The Averaging Masking Effect}
\label{sec:motivation}

Before elaborating on our proposed framework, we first analyze the limitations of standard dynamic pruning techniques~\citep{katharopoulos2018not,raju2021accelerating,qin2023infobatch} when applied to spatio-temporal domains. Existing methods typically quantify sample difficulty using a global loss metric, $\ell_i = \| f_\theta(X_i) - Y_i \|$, where a lower $\ell_i$ indicates an ``easy'' sample.
However, in spatio-temporal forecasting, the scalar loss $\ell_i$ is an aggregate metric averaged over $N$ spatial nodes and $T$ time steps. This aggregation introduces a critical pathology we term the \textit{Averaging Masking Effect}.We empirically demonstrate this phenomenon in Figure~\ref{fig:insight} (a), which visualizes the error landscapes of two distinct samples from a real-world traffic dataset:
\begin{itemize}[leftmargin=*,parsep=0pt]
    \item \textbf{Sample A (Global Noise):} The error is distributed relatively uniformly across the spatio-temporal field. The global hardness (mean MAE) is calculated as 17.05.
    \item \textbf{Sample B (Local Anomaly):} As highlighted by the red circle in Figure~\ref{fig:insight}(a), this sample contains critical localized failures (\eg, severe congestion spikes at specific hubs), while the remaining nodes exhibit low errors. Surprisingly, its global hardness is 17.16.
\end{itemize}
Despite the fundamental difference in structural information, standard pruning methods viewing only the global hardness ($\approx 17.1$) would treat these two samples as identical. Consequently, the informative Sample B, which drives the learning of local spatial dynamics, may risks being pruned as 
``easy'' data. This observation confirms that magnitude-based metrics fail to distinguish structural complexity from background noise. 
To address this, we propose \textit{Spatio-Temporal Complexity Scoring}, a metric designed to capture the \textit{non-uniformity} ($\sigma$) of the error distribution, ensuring that samples with high spatial heterogeneity (like Sample B) are preserved even if their global mean is low.

\subsubsection{Spatio-Temporal Complexity Scoring}
\label{sec:scoring}

Formally, we propose \model. As shown in Fig.~\ref{fig:overall}, during the forward propagation process, each sample retains a score related to its loss value. The average of these scores is set as the pruning threshold. In each epoch, a corresponding proportion of samples with low scores are pruned.

To operationalize the insight from Figure~\ref{fig:insight}(a), we propose a composite scoring function that explicitly penalizes performance heterogeneity. Let $\mathbf{E}_t^{(i)} \in \mathbb{R}^{N \times T}$ denote the absolute error matrix for the $i$-th sample at epoch $t$. We define the structural informativeness score $\mathcal{H}_t(i)$ as:
\begin{equation}
    \mathcal{H}_t(i) = \underbrace{\mu(\mathbf{E}_t^{(i)})}_{\text{Global Hardness}} + \lambda \cdot \left[ \underbrace{\sigma_{\text{space}}(\mathbf{E}_t^{(i)}) + \sigma_{\text{time}}(\mathbf{E}_t^{(i)})}_{\text{Spatio-Temporal Complexity}} \right],
\end{equation}
where $\mu(\cdot)$ denotes the global mean. The terms $\sigma_{\text{space}}$ and $\sigma_{\text{time}}$ represent the standard deviations calculated along the spatial and temporal dimensions, respectively. 

\subsubsection{Randomized Pruning Policy}
To reduce computational overhead while maintaining data diversity, we categorize samples into an \textit{Informative Set} $\mathcal{S}_{inf}$ and a \textit{Redundant Set} $\mathcal{S}_{red}$ at the beginning of each epoch $t$.
Let $\bar{\mathcal{H}}_t$ be the average score of the entire dataset.
\begin{equation}
    z_i \in \begin{cases}
        \mathcal{S}_{inf}, & \text{if } \mathcal{H}_t(i) \geq \bar{\mathcal{H}}_t \\
        \mathcal{S}_{red}, & \text{if } \mathcal{H}_t(i) < \bar{\mathcal{H}}_t
    \end{cases}
\end{equation}
Unlike static pruning which permanently discards $\mathcal{S}_{red}$, we adopt a randomized ``soft'' pruning strategy.
All samples in $\mathcal{S}_{inf}$ are retained.
For samples in $\mathcal{S}_{red}$, we retain them with a probability $p \in (0, 1)$, determined by the target pruning ratio. This ensures that even currently ''easy`` samples have a chance to be revisited, preventing the model from catastrophic forgetting of basic patterns.

\subsection{Stability-Guided Optimization}

\subsubsection{Stationarity-Aware Gradient Rescaling}
\label{sec:rescaling}

While the scoring metric identifies which samples to prune, a naive removal of ``easy'' samples would inevitably alter the training data distribution. As highlighted in Figure~\ref{fig:insight}(b), spatio-temporal datasets exhibit a long-tail distribution of dynamic intensity: the vast majority of samples are stationary (low temporal variance), while high-dynamic events are rare.
Standard pruning techniques~\cite{qin2023infobatch}, which apply a uniform rescaling weight $w = \frac{1}{1-r}$ (where $r$ is the pruning rate), fail to account for this imbalance. By disproportionately removing stationary samples, they shift the training distribution towards the ``tail" (non-stationary events), causing the model to overfit to extreme dynamics and lose robustness on regular patterns.

To rectify this distribution shift, we propose a \textit{Stationarity-Aware Rescaling} strategy. We first quantify the \textit{dynamic intensity} $\delta_i$ of each sample $i$ as the temporal variance of its ground truth targets: $\delta_i = \operatorname{Var}_{t}(Y_i)$.
For the subset of retained ``inf" samples $\mathcal{S}_{inf}'$, we assign an adaptive weight $w_i$ that is inversely proportional to their dynamic intensity:
\begin{equation}
    w_i = \frac{1}{1-r} \cdot \underbrace{\left( \frac{\bar{\delta}_{\mathcal{D}}}{\delta_i + \epsilon} \right)^\alpha}_{\text{Stationarity Correction}},
\end{equation}
where $\bar{\delta}_{\mathcal{D}}$ is the global average intensity, and $\alpha \ge 0$ controls the correction strength.
This mechanism ensures that preserved stationary samples ($\text{low } \delta_i$) receive higher weights, effectively representing the population of pruned stationary data. Conversely, highly dynamic samples ($\text{high } \delta_i$) have been naturally retained through our scoring metric and will receive standard weights.
Formally, the modified objective $\tilde{\mathcal{J}}(\theta)$ for the current epoch becomes:
\begin{equation}
    \tilde{\mathcal{J}}(\theta) = \sum_{i \in \mathcal{S}_{red}} \mathcal{L}(x_i, y_i; \theta) + \sum_{j \in \mathcal{S}_{inf}'} w_j \cdot \mathcal{L}(x_j, y_j; \theta).
\end{equation}
This reweighting guarantees that the gradient expectation remains unbiased not only in magnitude but also in terms of \textit{dynamic regime distribution}, ensuring convergence consistency with the full dataset.

\subsubsection{Training Schedule with Annealing}
While the unbiased estimator holds in expectation, the variance of the gradients inevitably increases due to downsampling. To mitigate this effect and ensure stable convergence in the final phase of training, we introduce a deterministic annealing strategy.
Given a total training budget of $E$ epochs and a cutoff ratio $\delta$ (\eg, $\delta=0.9$), we perform the pruning strategy only for the first $\delta \cdot E$ epochs. For the remaining epochs $t > \delta \cdot E$, we revert to full-dataset training. This allows the model to fine-tune on all samples, eliminating any residual variance and ensuring that the final model performance is strictly lossless compared to the baseline.

%% file: sections/05_experiments.tex
\section{Experiments}

In this section,we conduct extensive experiments to answer the following research questions (RQs):
\begin{itemize}[leftmargin=*,itemsep=0em]
    \item \textbf{RQ1:} Can \model outperform existing data pruning methods in various ST datasets? \textit{(\textbf{Effectiveness})}
    \item \textbf{RQ2:} How does the efficiency of \model compare with that of the existing baselines? \textit{(\textbf{Efficiency})}
    \item \textbf{RQ3:} How scalable is \model with respect to large-scale datasets and foundation models? \textit{(\textbf{Scalability})}
    \item \textbf{RQ4:} Is \model effective across different types of backbones, optimizers, and tasks? \textit{(\textbf{Universality})}
    \item \textbf{RQ5:} How does \model work? Which components play the key roles, and are they sensitive to parameter settings or component designs? \textit{(\textbf{Mechanism \& Robustness})}
\end{itemize}

\subsection{Experimental Setup}

\textbf{Dataset and Evaluation Protocol.}
We evaluate our method on representative ST benchmarks: \PemsEight~\citep{song2020spatial} from the traffic domain and \UrbanEV~\citep{li2025urbanev} from the energy domain, alongside the large-scale benchmark \LargeST~\citep{liu2024largest} for scalability analysis. Following standard protocols~\citep{shao2023exploring,jin2023spatio}, datasets are split chronologically (6:2:2) to forecast the next 12 steps given the past 12. To ensure rigorous efficiency comparison, we disable early stopping and train for a fixed 100 epochs. Each experiment was repeated five times and the mean was reported. Performance under different pruning levels is assessed via MAE, RMSE, and MAPE across data retention ratios of \{10\%, 30\%, 50\%, 70\%\}. Further details of datasets and protocols in Appendix~\ref{appendix_datasets}.

\textbf{Baselines and Parameter Settings.}
We benchmark against a comprehensive suite of pruning strategies adapted for spatio-temporal tasks, categorized into: (1) \textit{Static methods}: Random (\HardRandom), geometry-based (\CD, \Herding, \Kmeans), uncertainty-based (\LeastConfidence, \Entropy, \Margin), loss-based (\Forgetting, \GraNd), decision boundary-based (\Cal), bi-level (\Glister), and submodular (\GraphCut, \FaLo); and (2) \textit{Dynamic methods}: Random(\SoftRandom), uncertainty-based (\ThetaGreedy, \UCB), and loss-based (\InfoBatch).
We employ \GWNet~\citep{wu2019graph} as the default backbone, alongside \STAEformer~\citep{liu2023spatio} and \STID~\citep{shao2022spatial} for cross-architecture evaluation, and the foundation model \Opencity~\citep{Li2025OpenSF} to assess scalability. To ensure fair comparison, all models are trained using SGD (momentum 0.9, weight decay 1e-4) with a cosine annealing scheduler. More details on baselines, models, and parameters are provided in Appendix~\ref{appendix_baseline}.

\subsection{Effectiveness Study (RQ1)}

Table \ref{tab:rq1_mape} compares the MAPE results of \model against representative data pruning and selection methods. The best results are highlighted in \firstres{bold pink}, and the second-best in \secondres{underlined in blue}. Besides, we report the relative performance change (\blue{degradation} or \red{improvement}) compared to the whole dataset baseline. We observe that: \ding{182} \textit{Dynamic pruning strategies generally outperform static ones.} Across various baselines, dynamic methods achieve superior performance by leveraging training dynamics to better assess sample importance. Interestingly, both Hard Random and Soft Random remain competitive at low retention rates, mirroring phenomena observed in vision datasets~\citep{guo2022deepcore}. We hypothesize that for strategy-based methods, specific algorithmic inductive biases may introduce a harmful sampling expectation bias. \ding{183} \textit{Dataset redundancy varies across domains.} We observe that \PemsEight~is more sensitive to pruning than \UrbanEV, indicating differing redundancy levels. For example, at 10\% retention (excluding our method), performance degradation on \PemsEight~($20.40\% \sim 45.26\%$) significantly exceeds that of \UrbanEV~($7.34\% \sim 28.63\%$). However, performance recovers rapidly as the retention ratio increases; at a 50\% pruning level, degradation across both datasets remains within single digits. \ding{184} \textit{\model consistently excels across all pruning ratios.} \model outperforms both static and dynamic baselines in all settings. Even at 10\% retention rate, it keeps practical utility with minimal degradation. Notably, \model surpasses whole dataset performance on \UrbanEV. We attribute this to \UrbanEV's low redundancy and high signal-to-noise ratio, where our method effectively filters noise to boost performance.

\input{tables/rq1_table}

\input{tables/rq2_figure}

\subsection{Efficiency Study (RQ2)}

Figure \ref{fig:rq2_efficiency} illustrates the efficiency comparison. We observe the following: \ding{185} \textit{\model strikes a superior balance between training time and performance,} significantly outperforming other baselines. Notably, \model achieves nearly $2\times$ acceleration ($\approx 50\%$ reduction in per-epoch time) with negligible performance loss across different metrics. \ding{185} Even under an aggressive $10\times$ speedup, \model incurs only marginal degradation, maintaining acceptable performance where competing methods deteriorate rapidly.

\subsection{Scalability Study (RQ3)}

To assess the scalability of our method, we conduct evaluations from both data and model perspectives. Specifically, we validate performance using the representative large-scale spatio-temporal dataset \LargeST~\citep{liu2024largest}, which comprises the SD, GBA, and GLA subsets. Additionally, we employ the spatio-temporal foundation model \Opencity~\citep{Li2025OpenSF} across the Mini, Base, and Plus scales.

\input{tables/rq3_table}

\textbf{Large-Scale Datasets.}
Table \ref{tab:rq3_comparison} presents comparative performance on the LargeST benchmark. We highlight the following observations: \ding{182} \textit{Competitive efficiency-performance trade-off.} At a 10\% data retention rate, heuristic baselines (\ie, \SoftRandom) achieve the fastest dynamic pruning speeds but suffer significant performance degradation. In contrast, our method achieves relative performance gains of $11.09\%\sim38.80\%$ over heuristics and even slightly outperforms full-data training, despite the additional computational overhead incurred by our customized policy design. \ding{183} \textit{Robustness across spatio-temporal scales.} Unlike heuristic baselines (\ie, \SoftRandom) that exhibit marked performance drops on smaller datasets (\eg, SD), our \model maintains consistent performance advantages across varying spatio-temporal scales. \ding{184} \textit{Superior efficiency and scalability.} Even in extreme 1\% retention scenarios, our model retains competitive performance while drastically reducing computational costs. By cutting training time from days to hours, it demonstrates significant potential for data scalability and large-scale deployment.

\input{tables/rq3_figure}

\textbf{ST-Foundation Models.}
Following \citep{Li2025OpenSF} settings, we further validate model scalability by integrating \model with the \Opencity~foundation model series. Figure~\ref{fig:rq3_stfm} illustrates the trade-off between per-epoch pre-training time and downstream performance, revealing the following: \ding{182} \textit{Simultaneous gains in efficiency and capability.} \model achieves a ``win-win'' outcome, with green arrow consistently pointing towards the ``most sample-efficient'' region. Across all model scales, \model not only significantly reduces the pre-training time (shifting left) but also generally improves the prediction accuracy (shifting upwards). This suggests that our approach effectively purifies the pre-training corpus by eliminating noisy or redundant samples that hinder convergence. \ding{183} \textit{Democratizing large-scale model training.} For the computation-intensive Plus scale, \model significantly curtails per-epoch training time (\eg, from $\sim$350 to 250 mins) while outperforming the original baseline. Notably, \model reduces the training cost of the Base scale to a level smaller than the Mini scale,  thus effectively training larger and more powerful models under constrained computational resources.

\subsection{Universality Study (RQ4)}

\input{tables/rq4_figure}

To evaluate the universality of our method, we conduct evaluations on \UrbanEV~across different backbones, optimizers, and tasks. Specifically, we validate performance using representative ST architectures including GWNet, STID, and STAEformer; distinct optimizers such as SGD, Adam~\citep{KingmaB15adam}, and Muon~\citep{jordan2024muon}; and prediction horizons covering short-term (\ie, 12$\rightarrow$12), medium-term (\ie, 24$\rightarrow$24), and long-term (\ie, 96$\rightarrow$96) tasks. From each subgraph in Figure ~\ref{fig:rq4_universality}, we observed that:

\textbf{Cross-Architecture Evaluation.} 
\ding{182} \textit{Universal performance gains.} Notably, across all three backbones, models trained on \model-selected subsets (dashed lines) consistently outperform their full-dataset counterparts (solid lines), achieving lower MAPE at most retention levels. \ding{183} \textit{Architectural robustness.} In low-retention regimes, our method yields pronounced gains for lightweight architectures like STID and GWNet. However, even for the advanced, over-parameterized STAEformer, our model can also push performance boundaries as the retention ratio increases.

\textbf{Cross-Optimizer Evaluation.} 
\ding{182} \textit{Robust adaptability to optimization process.} \model maintains effectiveness regardless of the chosen optimizer. Notably, with SGD, \model achieves substantial gains, outperforming the full-dataset baseline even at low retention rates. \ding{183} \textit{Enhancing advanced optimizers.} Even with state-of-the-art optimizers like Adam and the recent Muon, which already yield low error rates ($<24\%$), \model further pushes performance boundaries. At retention rates above $30\%$, \model successfully filters detrimental noise, enabling both optimizers to surpass their full-dataset baselines, thereby confirming the robustness of our method \model across diverse optimization landscapes.

\textbf{Cross-Task Evaluation.} 
\ding{182} \textit{Short-to-medium term robustness.} For short and medium horizons, \model consistently outperforms full-dataset baselines across all retention ratios, confirming that immediate temporal dependencies are effectively captured within compact, denoised subsets. \ding{183} \textit{Long-term sensitivity.} Although long-term involves more complex dependencies and higher sensitivity to data scarcity, our method remains robust; performance successfully recovers to match full-dataset levels at a 50\% retention ratio.

\subsection{Mechanism \& Robustness Study (RQ5)}

\textbf{Ablation Study.}
To validate the effectiveness of each component in \model, we compare against four variants: removing the spatio-temporal complexity score (\textit{w/o STC}), the score rescaling strategy (\textit{w/o Res.}), and the annealing scheduler (\textit{w/o Anne.}). As shown in Figure \ref{fig:rq5_mechanism} Left, \model consistently yields the lowest errors across all retention ratios. We observe that: \ding{182} \textit{Necessity of scheduling with annealing.} The performance degradation of \textit{w/o Anne.} is the most significant, especially under low retention rates, which illustrates its positive role in variance correction for sparse scenarios. \ding{183} \textit{Importance of Scoring Components.} Removing \textit{STC} leads to noticeable error increases, indicating that both the intrinsic difficulty of samples and the historical consistency are critical for robust sample valuation.

\input{tables/rq5_figure_mechanism}

\textbf{Parameter Sensitivity.}
We further investigate the sensitivity of \model to key hyperparameters: the complexity weight  and the annealing rate .
\ding{182} \textit{Impact of Complexity Weight .} Figure \ref{fig:rq5_mechanism} (Middle) exhibits a convex trend for , where extreme values (too small or too large) hamper performance. The optimal range lies between 0.4 and 0.6, suggesting that balancing the raw loss with spatio-temporal structural complexity is essential for identifying informative samples.
\ding{183} \textit{Impact of Annealing Rate .} Figure \ref{fig:rq5_mechanism} (Right) reveals that performance is stable when $\delta\in[0.8,0.95]$ but deteriorates rapidly as $\delta$ approaches $1$, \ie~\textit{w/o Anne.}, consistent with previous observations in the ablation study.

\input{tables/rq5_figure_cluster}

\textbf{Interpretability Study.}
To intuitively understand how \model selects spatio-temporal sample compared to baselines, we visualize the latent distribution of the selected subsets using t-SNE in Figure \ref{fig:rq5_cluster}. We can observe that: \ding{182} \textit{Dynamic alignment with training dynamics.} While heuristic baselines (\eg, \textit{Hard/Soft Random}) produce static, sparse distributions via uniform sampling, \model exhibits a clear evolutionary trajectory, balancing the preservation of representative centroids with the exploration of global diversity. \ding{183} \textit{Restoration of intrinsic data topology.} Surpassing \InfoBatch, \model rapidly reconstructs the approximate manifold topology of the original distribution during training, thereby ensuring robust generalization performance.

%% file: tables/rq1_table.tex
\begin{table*}[t!]
    \centering
    \caption{Performance comparison to state-of-the-art dataset selection and pruning methods when remaining $\{10\%,30\%,50\%,70\%\}$ of the full training set. All methods are trained using same ST backbone. (Additional MAE and RMSE results are provided in Appendix \ref{appendix_moreresult})}
    \label{tab:rq1_mape}
    \footnotesize
    \renewcommand{\arraystretch}{1.4}
    \setlength{\tabcolsep}{3pt}
    \resizebox{\textwidth}{!}{
    \begin{tabular}{cc|cccc|cccc}
    \toprule
    \multirow{3}{*}{} & Dataset  & \multicolumn{4}{c|}{\textsc{Pems08} ( MAPE (\%) $\downarrow$)} & \multicolumn{4}{c}{\textsc{UrbanEV} ( MAPE (\%) $\downarrow$)}  \\
    \midrule
    & Remaining Ratio \% & 10 & 30 & 50 & 70  & 10 & 30 & 50 & 70 \\ \midrule
    \parbox[t]{4mm}{\multirow{13}{*}{\rotatebox[origin=c]{90}{Static}}}
    
    & Hard Random
    & 13.93\blue{26.87\%} & 14.80\blue{34.79\%} & 12.07\blue{9.93\%} & 12.06\blue{9.84\%}
    & 35.89\blue{20.32\%} & 31.90\blue{6.94\%} & 31.52\blue{5.67\%} & 30.69\blue{2.88\%}
    \\
    
    & CD~\citep{agarwal2020contextual}
    & 14.55\blue{32.51\%} & 12.25\blue{11.57\%} & 11.96\blue{8.93\%} & 11.83\blue{7.74\%}
    & 38.37\blue{28.63\%} & 32.20\blue{7.95\%} & 30.40\blue{1.91\%} & 31.01\blue{3.96\%} \\
    
    & Herding~\citep{welling2009herding}
    & 15.79\blue{43.81\%} & \secondres{12.11}\blue{10.29\%} & 11.58\blue{5.46\%} & 12.84\blue{16.94\%}
    & 34.63\blue{16.09\%} & 30.97\blue{3.82\%} & 30.94\blue{3.72\%} & 30.97\blue{3.82\%}
    \\
    
    & K-Means~\citep{sener2018active}
    & 15.71\blue{43.08\%} & 12.24\blue{11.48\%} & 12.48\blue{13.66\%} & 11.52\blue{4.92\%}
    & 38.45\blue{28.90\%} & 31.85\blue{6.77\%} & 30.96\blue{3.79\%} & 30.69\blue{2.88\%}
    \\
    
    & Least Confidence~\citep{coleman2019selection}
    & 14.31\blue{30.33\%} & 12.62\blue{14.94\%} & 11.74\blue{6.92\%} & 11.61\blue{5.74\%}
    & 34.23\blue{14.75\%} & 36.87\blue{23.60\%} & 30.98\blue{3.86\%} & 30.55\blue{2.41\%}
    \\
    
    & Entropy~\citep{coleman2019selection}
    & \secondres{13.22}\blue{20.40\%} & 12.82\blue{16.76\%} & 12.83\blue{16.85\%} & 11.28\blue{2.73\%}
    & 36.42\blue{22.09\%} & 31.20\blue{4.59\%} & 30.65\blue{2.75\%} & 30.49\blue{2.21\%}
    \\
    
    & Margin~\citep{coleman2019selection}
    & 14.06\blue{28.05\%} & 12.55\blue{14.30\%} & 11.60\blue{5.65\%} & 11.38\blue{3.64\%}
    & 37.21\blue{24.74\%} & 31.89\blue{6.91\%} & 32.73\blue{9.72\%} & 30.89\blue{3.55\%}
    \\
    
    & Forgetting~\citep{toneva2018empirical}
    & 14.92\blue{35.88\%} & 12.19\blue{11.02\%} & \secondres{11.46}\blue{4.37\%} & 11.29\blue{2.82\%}
    & 32.84\blue{10.09\%} & 31.36\blue{5.13\%} & 30.63\blue{2.68\%} & 30.69\blue{2.88\%}
    \\
    
    & GraNd~\citep{paul2021deep}
    & 13.93\blue{26.87\%} & 13.09\blue{19.22\%} & 12.33\blue{12.30\%} & 11.50\blue{4.74\%}
    & 35.26\blue{18.20\%} & 31.71\blue{6.30\%} & 30.90\blue{3.59\%} & 30.20\blue{1.24\%}
    \\
    
    & Cal~\citep{margatina2021active}
    & 14.79\blue{34.70\%} & 12.40\blue{12.93\%} & 12.64\blue{15.12\%} & 12.40\blue{12.93\%}
    & 33.77\blue{13.21\%} & 32.25\blue{8.11\%} & 31.45\blue{5.43\%} & 30.81\blue{3.29\%}
    \\
    
    & Glister~\citep{killamsetty2021glister}
    & 14.48\blue{31.88\%} & 13.23\blue{20.49\%} & 12.51\blue{13.93\%} & 11.74\blue{6.92\%}
    & 33.70\blue{12.97\%} & 33.65\blue{12.81\%} & 32.78\blue{9.89\%} & 31.26\blue{4.79\%} \\
    
    & GraphCut~\citep{iyer2021submodular}
    & 15.07\blue{37.25\%} & 12.26\blue{11.66\%} & 11.84\blue{7.83\%} & 11.27\blue{2.64\%}
    & 35.73\blue{19.78\%} & 31.49\blue{5.56\%} & 30.76\blue{3.12\%} & 31.23\blue{4.69\%}
    \\
    
    & FaLo~\citep{iyer2021submodular}
    & 15.95\blue{45.26\%} & 12.56\blue{14.39\%} & 12.13\blue{10.47\%} & 11.40\blue{3.83\%}
    & 35.07\blue{17.57\%} & 31.72\blue{6.34\%} & 31.23\blue{4.69\%} & 31.01\blue{3.96\%}
    \\
    
    \midrule
    \parbox[t]{4mm}{\multirow{5}{*}{\rotatebox[origin=c]{90}{Dynamic}}}
    & Soft Random
    & 15.00\blue{36.61\%} & 12.41\blue{13.02\%} & 12.01\blue{9.38\%} & 11.34\blue{3.28\%}
    & 33.55\blue{12.47\%} & 31.34\blue{5.06\%} & 30.48\blue{2.18\%} & 30.35\blue{1.74\%}
    \\

    & $\epsilon$-greedy~\citep{raju2021accelerating}
    & 13.82\blue{25.87\%} & 12.21\blue{11.20\%} & 11.84\blue{7.83\%} & \secondres{11.24}\blue{2.37\%}
    & 33.25\blue{11.46\%} & 30.98\blue{3.86\%} & 30.46\blue{2.11\%} & 30.30\blue{1.58\%}
    \\

    & UCB~\citep{raju2021accelerating}
    & 13.80\blue{25.68\%} & 12.41\blue{13.02\%} & 11.76\blue{7.10\%} & 11.28\blue{2.73\%}
    & 33.24\blue{11.43\%} & 31.08\blue{4.19\%} & 30.43\blue{2.01\%} & 30.29\blue{1.54\%}
    \\

    & InfoBatch~\citep{qin2023infobatch}
    & 13.66\blue{24.41\%} & 12.12\blue{10.38\%} & 11.83\blue{7.74\%} & 11.33\blue{3.19\%}
    & \secondres{32.02}\blue{7.34\%} & \secondres{30.78}\blue{3.18\%} & \secondres{30.24}\blue{1.37\%} & \secondres{29.83}\blue{0.00\%}
    \\

    & \model (\textcolor{orange}{Our})
    & \colorbox[HTML]{DAE8FC}{\firstres{12.13}\blue{\textbf{13.21}}} 
    & \colorbox[HTML]{DAE8FC}{\firstres{11.75}\blue{\textbf{7.01}}} 
    & \colorbox[HTML]{DAE8FC}{\firstres{11.11}\blue{\textbf{5.56}}} 
    & \colorbox[HTML]{DAE8FC}{\firstres{11.19}\blue{\textbf{2.82}}}

    & \colorbox[HTML]{DAE8FC}{\firstres{29.56}\red{\textbf{0.91}}}
    & \colorbox[HTML]{DAE8FC}{\firstres{29.42}\red{\textbf{1.37}}} 
    & \colorbox[HTML]{DAE8FC}{\firstres{29.15}\red{\textbf{2.28}}} 
    & \colorbox[HTML]{DAE8FC}{\firstres{29.05}\red{\textbf{2.61}}}

    \\

    \midrule
    \multicolumn{2}{c|}{Whole Dataset} & \multicolumn{4}{c|}{10.98$_{\pm0.23}$} & \multicolumn{4}{c}{29.83$_{\pm0.09}$} \\
    \bottomrule
    \end{tabular}}
\end{table*}

%% file: tables/rq2_figure.tex
\begin{figure*}[t!]
\centering
\begin{minipage}[t]{0.32\linewidth}
  \centering
  \includegraphics[width=\linewidth]{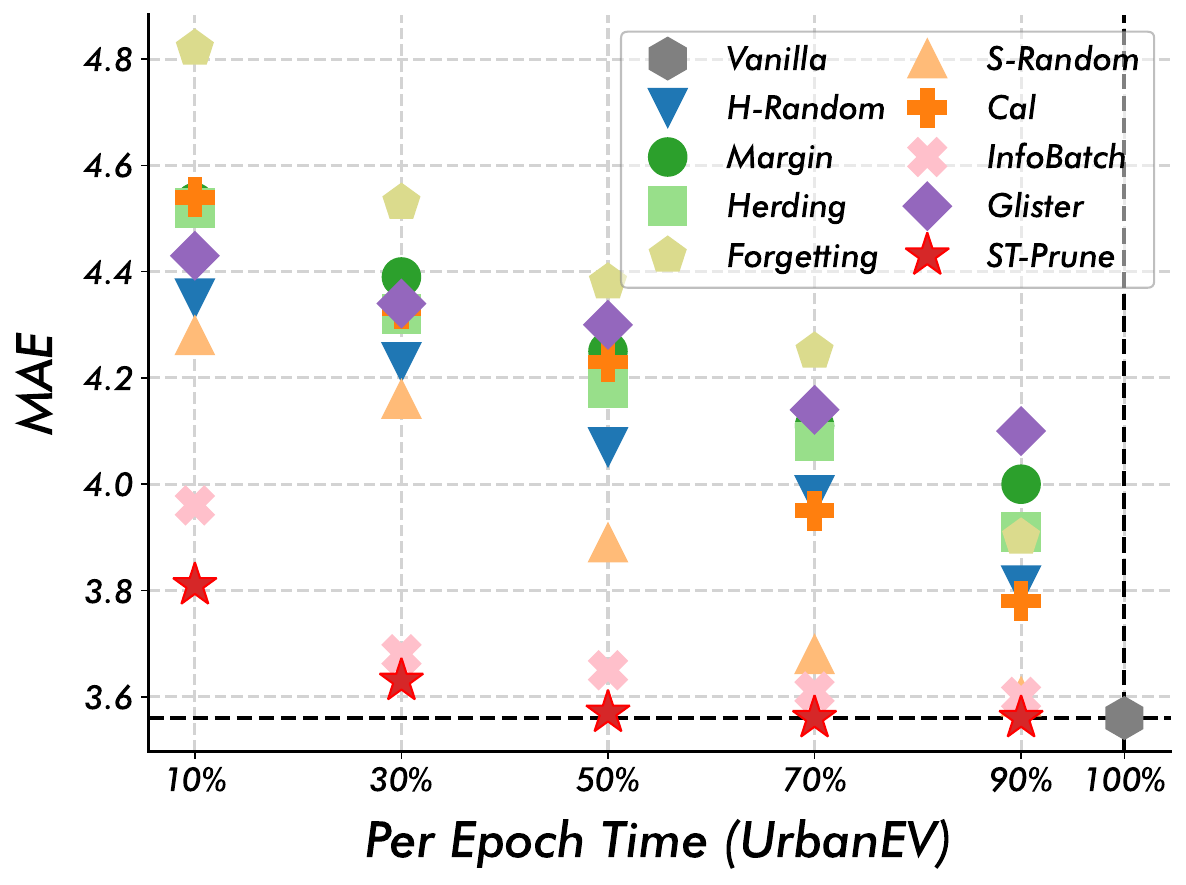}
\end{minipage}\hfill
\begin{minipage}[t]{0.32\linewidth}
  \centering
  \includegraphics[width=\linewidth]{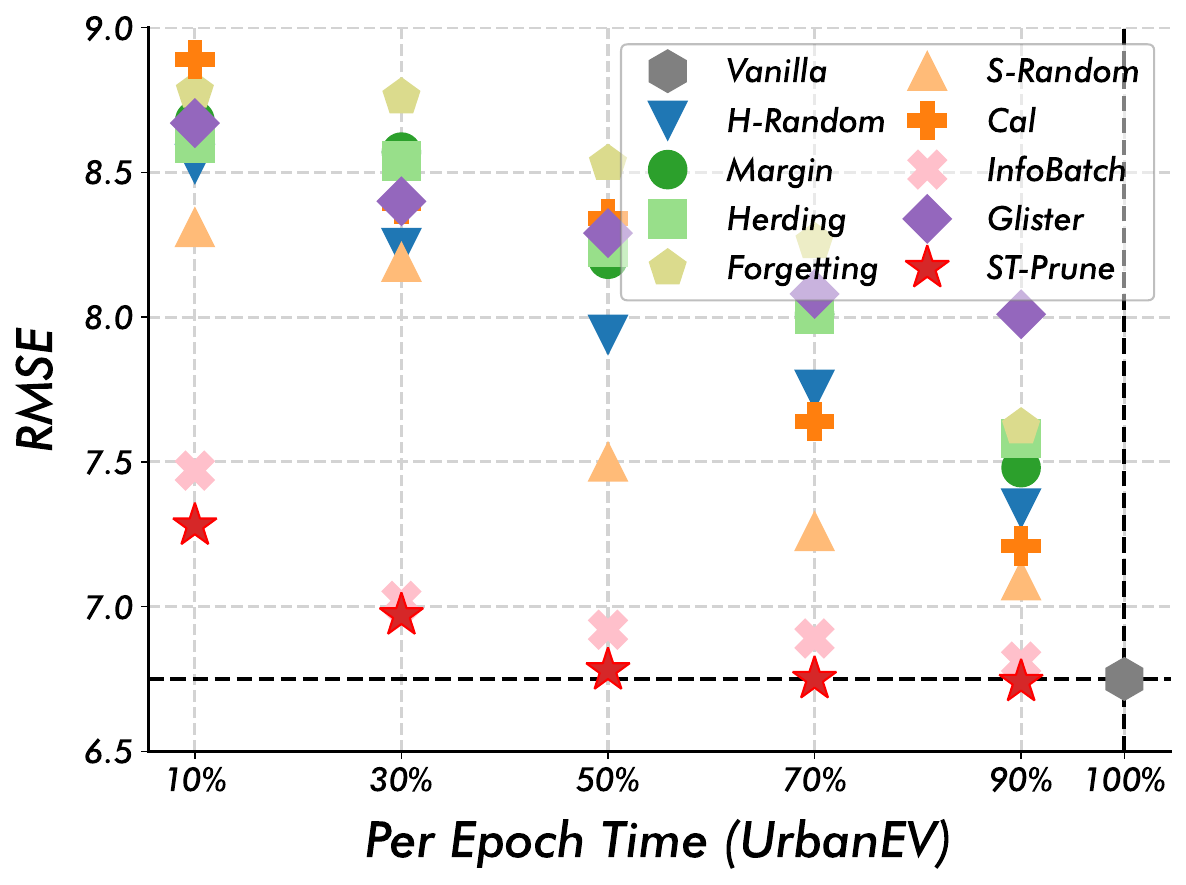}
\end{minipage}\hfill
\begin{minipage}[t]{0.32\linewidth}
  \centering
  \includegraphics[width=\linewidth]{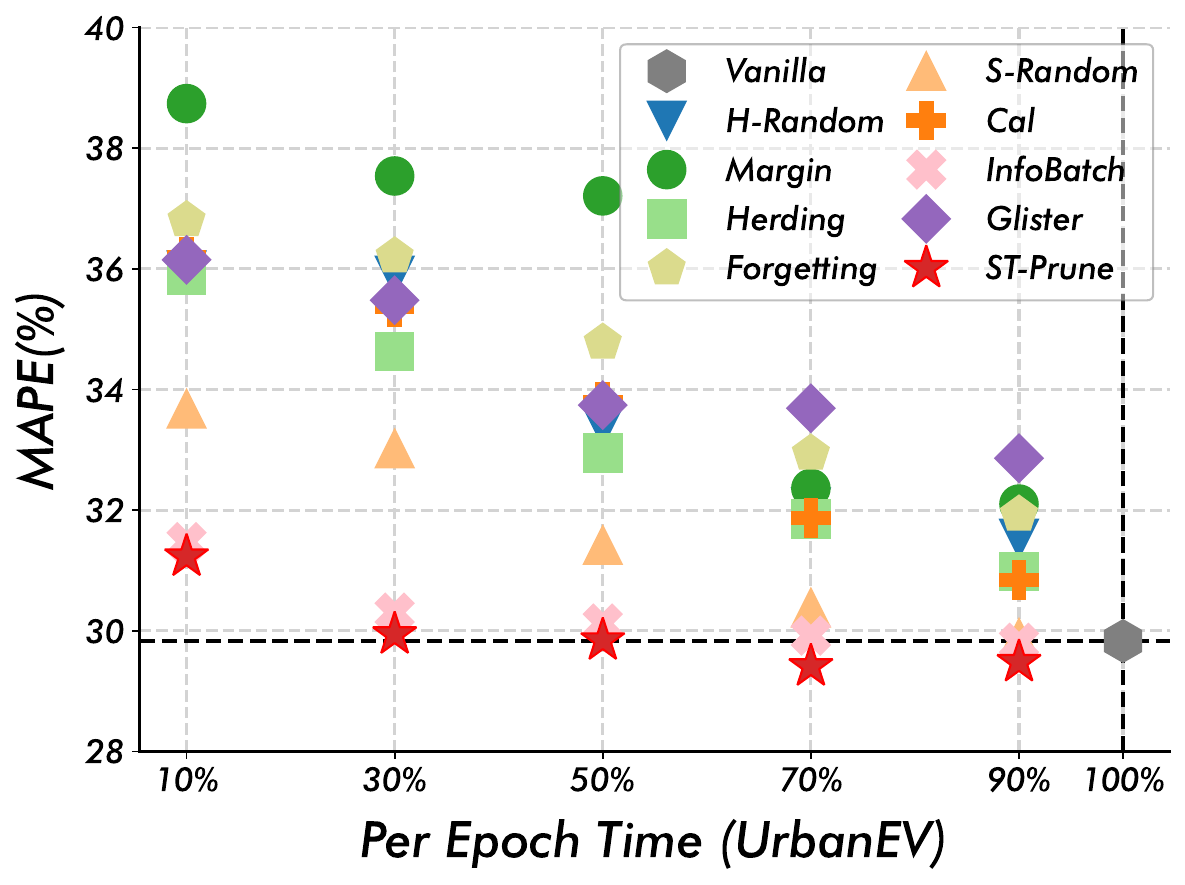}
\end{minipage}
\caption{The trade-off between per-epoch time and performance in \UrbanEV. Specifically, we report the test performance when methods achieve per epoch times of \{10\%, 30\%, 50\%, 70\%, 90\%\} of the full dataset training time. ``Vanilla'' denotes the full dataset training result.}
\label{fig:rq2_efficiency}
\end{figure*}

%% file: tables/rq3_table.tex
\begin{table*}[htbp!]
\centering
\caption{Comparison of \model with the fastest dynamic pruning method \SoftRandom~on \LargeST. TIME: Total wall-clock time.}
\label{tab:rq3_comparison}
\renewcommand{\arraystretch}{1.6}
\setlength{\tabcolsep}{2pt}
\resizebox{\textwidth}{!}{
\begin{tabular}{c|cccc|cccc|cccc}
\toprule
\multirow{2}{*}{Dataset} 
& \multicolumn{4}{c}{SD ($\# 716$)} 
& \multicolumn{4}{c}{GBA ($\# 2352$)} 
& \multicolumn{4}{c}{GLA ($\# 3834$)} \\
\cline{2-13}
& MAE & RMSE & MAPE (\%) & TIME 
& MAE & RMSE & MAPE (\%) & TIME 
& MAE & RMSE & MAPE (\%) & TIME \\
\hline
Whole Dataset 
& 24.13 $_{\pm1.14}$ & 38.03 $_{\pm1.32}$ & 15.77 $_{\pm0.87}$ & \colorbox[HTML]{dafce3}{2.07h}
& 27.34 $_{\pm1.04}$ & 42.11 $_{\pm1.19}$ & 24.11 $_{\pm0.48}$ & \colorbox[HTML]{dafce3}{13.05h}
& 26.26 $_{\pm0.84}$ & 41.12 $_{\pm1.47}$ & 16.06 $_{\pm0.63}$ & \colorbox[HTML]{dafce3}{30.53h} \\
\hline
\SoftRandom~(10\%)
& \colorbox[HTML]{daf3fc}{30.41} & \colorbox[HTML]{daf3fc}{47.96} & \colorbox[HTML]{daf3fc}{20.43} & 0.52h 
& \colorbox[HTML]{daf3fc}{33.05} & \colorbox[HTML]{daf3fc}{48.71} & \colorbox[HTML]{daf3fc}{39.30} & 2.97h 
& \colorbox[HTML]{daf3fc}{29.74} & \colorbox[HTML]{daf3fc}{45.98} & \colorbox[HTML]{daf3fc}{20.97} & 7.25h \\
\hline
\model~(10\%)
& \colorbox[HTML]{daf3fc}{24.01} & \colorbox[HTML]{daf3fc}{37.73} & \colorbox[HTML]{daf3fc}{15.51} & {0.61h}
& \colorbox[HTML]{daf3fc}{27.32} & \colorbox[HTML]{daf3fc}{41.94} & \colorbox[HTML]{daf3fc}{24.05}& {4.92h}
& \colorbox[HTML]{daf3fc}{26.22} & \colorbox[HTML]{daf3fc}{40.88} & \colorbox[HTML]{daf3fc}{15.95} & {13.38h} \\

\model~(1\%)

& {24.76} & {38.19} & {16.51} & \colorbox[HTML]{dafce3}{0.31h}
& {28.16} & {42.77} & {23.64}& \colorbox[HTML]{dafce3}{2.84h}
& {26.95} & {41.57} & {17.62} & \colorbox[HTML]{dafce3}{7.12h} \\

\hline
Improvement
& \colorbox[HTML]{daf3fc}{$\uparrow 21.05\%$} & \colorbox[HTML]{daf3fc}{$\uparrow 21.33\%$} & \colorbox[HTML]{daf3fc}{$\uparrow 24.08\%$} & \colorbox[HTML]{dafce3}{$\times 6.68$}
& \colorbox[HTML]{daf3fc}{$\uparrow 17.33\%$} & \colorbox[HTML]{daf3fc}{$\uparrow 13.90\%$} & \colorbox[HTML]{daf3fc}{$\uparrow 38.80\%$} & \colorbox[HTML]{dafce3}{$\times 4.60$}
& \colorbox[HTML]{daf3fc}{$\uparrow 11.84\%$} & \colorbox[HTML]{daf3fc}{$\uparrow 11.09\%$} & \colorbox[HTML]{daf3fc}{$\uparrow 23.94\%$} & \colorbox[HTML]{dafce3}{$\times 4.29$} \\
\bottomrule
\end{tabular}
}
\end{table*}

%% file: tables/rq3_figure.tex

\begin{figure*}[htbp!]
    \centering
    \includegraphics[width=1.0\linewidth]{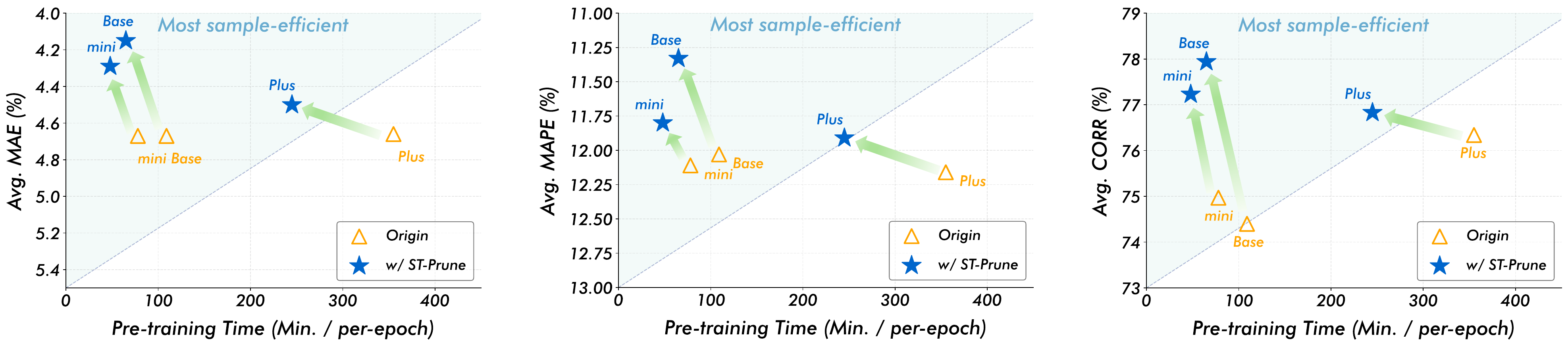}
    \caption{Performance and efficiency trade-offs between w/o and w/ \model at different scales of the ST-foundation model \Opencity.}
    \label{fig:rq3_stfm}
\end{figure*}

%% file: tables/rq4_figure.tex
\begin{figure*}[htbp!]
\centering
\begin{minipage}[t]{0.32\linewidth}
  \centering
  \includegraphics[width=\linewidth]{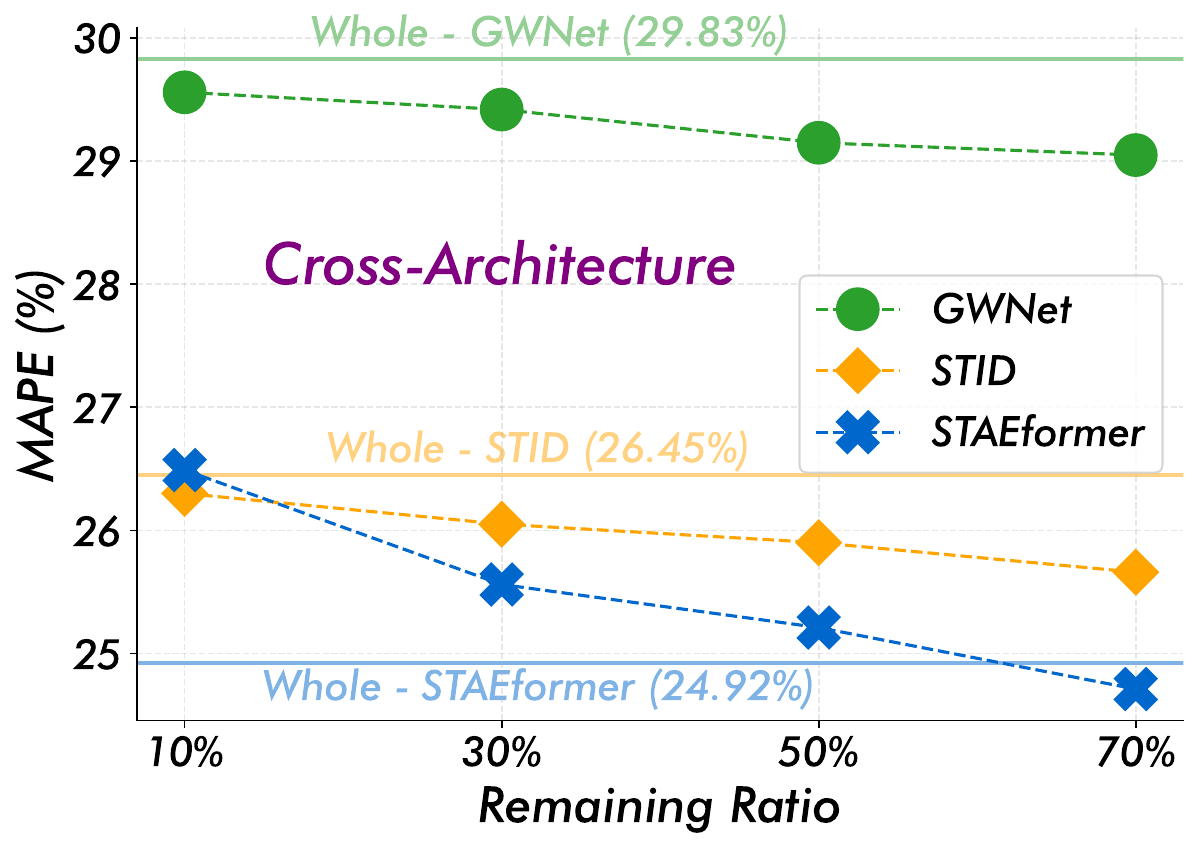}
\end{minipage}\hfill
\begin{minipage}[t]{0.32\linewidth}
  \centering
  \includegraphics[width=\linewidth]{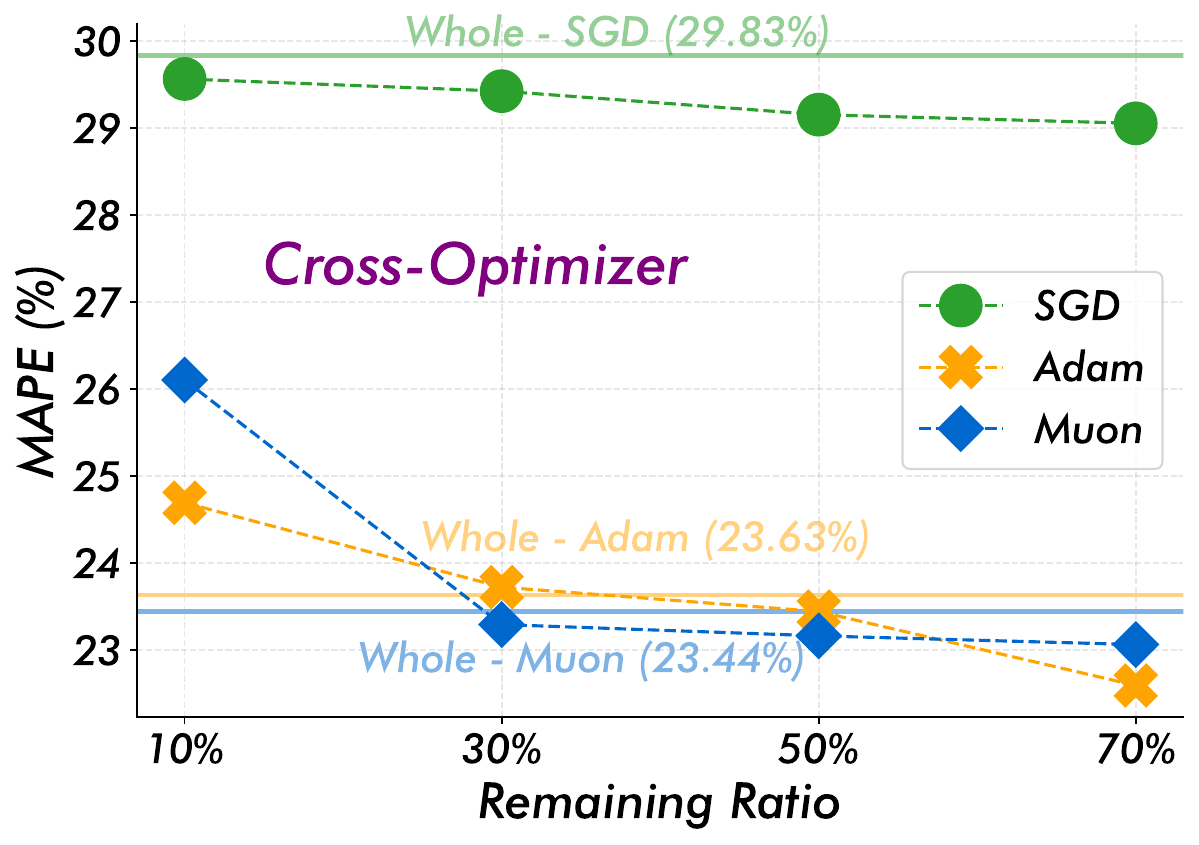}
\end{minipage}\hfill
\begin{minipage}[t]{0.32\linewidth}
  \centering
  \includegraphics[width=\linewidth]{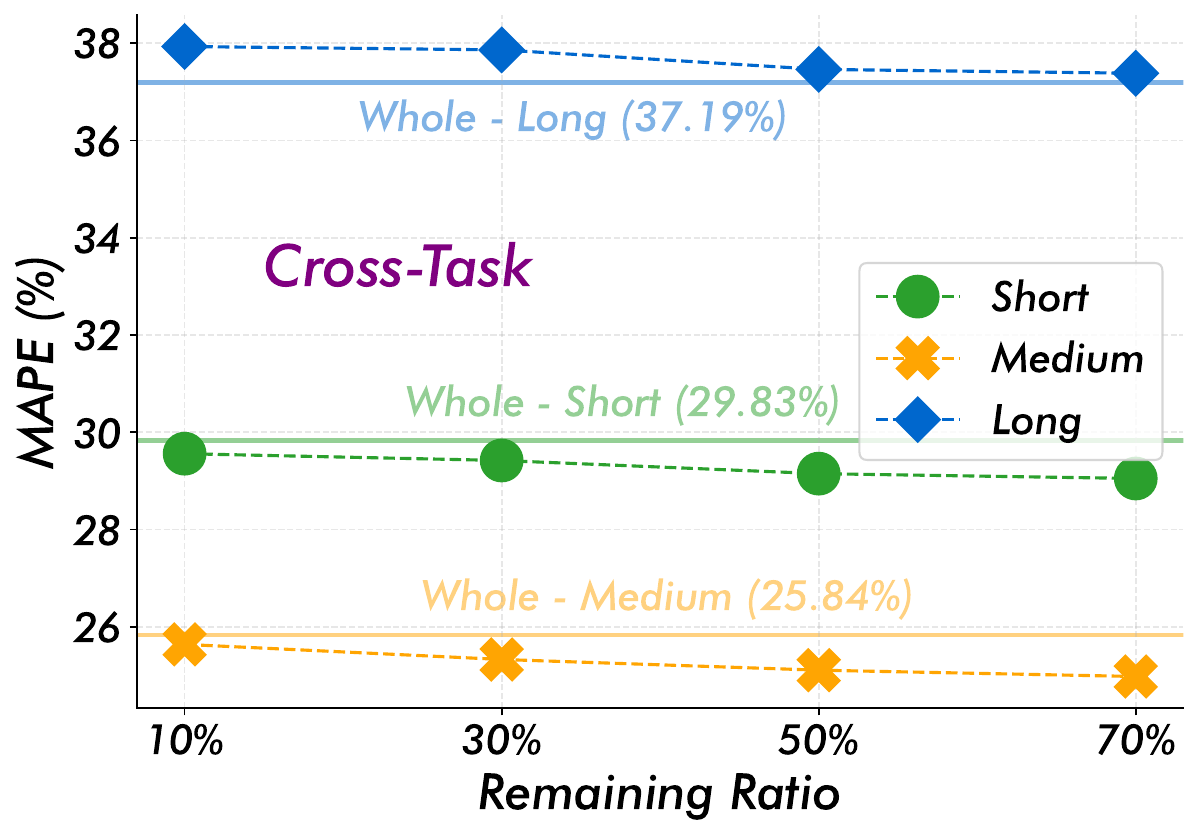}
\end{minipage}
\caption{Performance variations across distinct architectures, optimizers, and tasks under data retention rates of $\{10\%,30\%,50\%,70\%\}$.}
\label{fig:rq4_universality}
\end{figure*}

%% file: tables/rq5_figure_mechanism.tex
\begin{figure*}[htbp!]
\centering
\begin{minipage}[t]{0.32\linewidth}
  \centering
  \includegraphics[width=\linewidth]{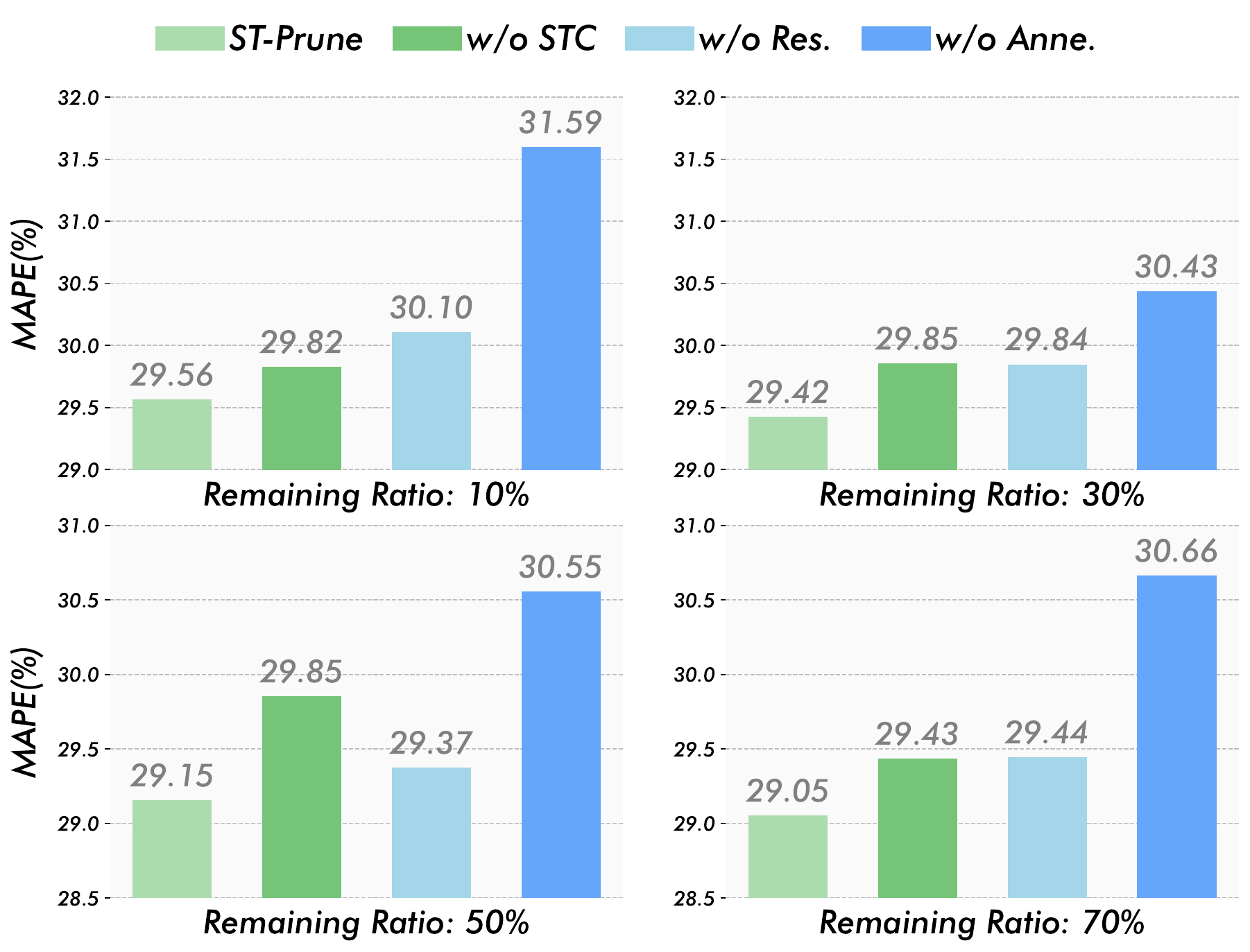}
\end{minipage}\hfill
\begin{minipage}[t]{0.32\linewidth}
  \centering
  \includegraphics[width=\linewidth]{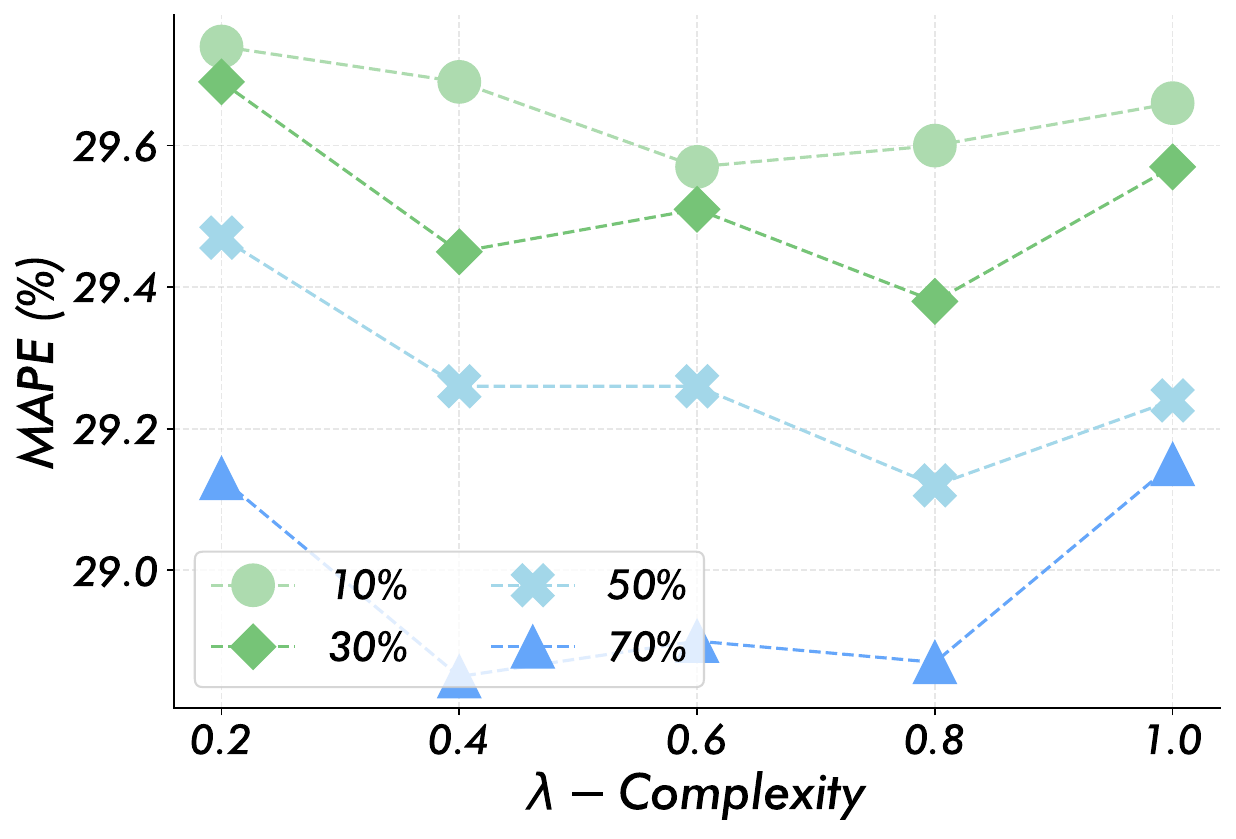}
\end{minipage}\hfill
\begin{minipage}[t]{0.32\linewidth}
  \centering
  \includegraphics[width=\linewidth]{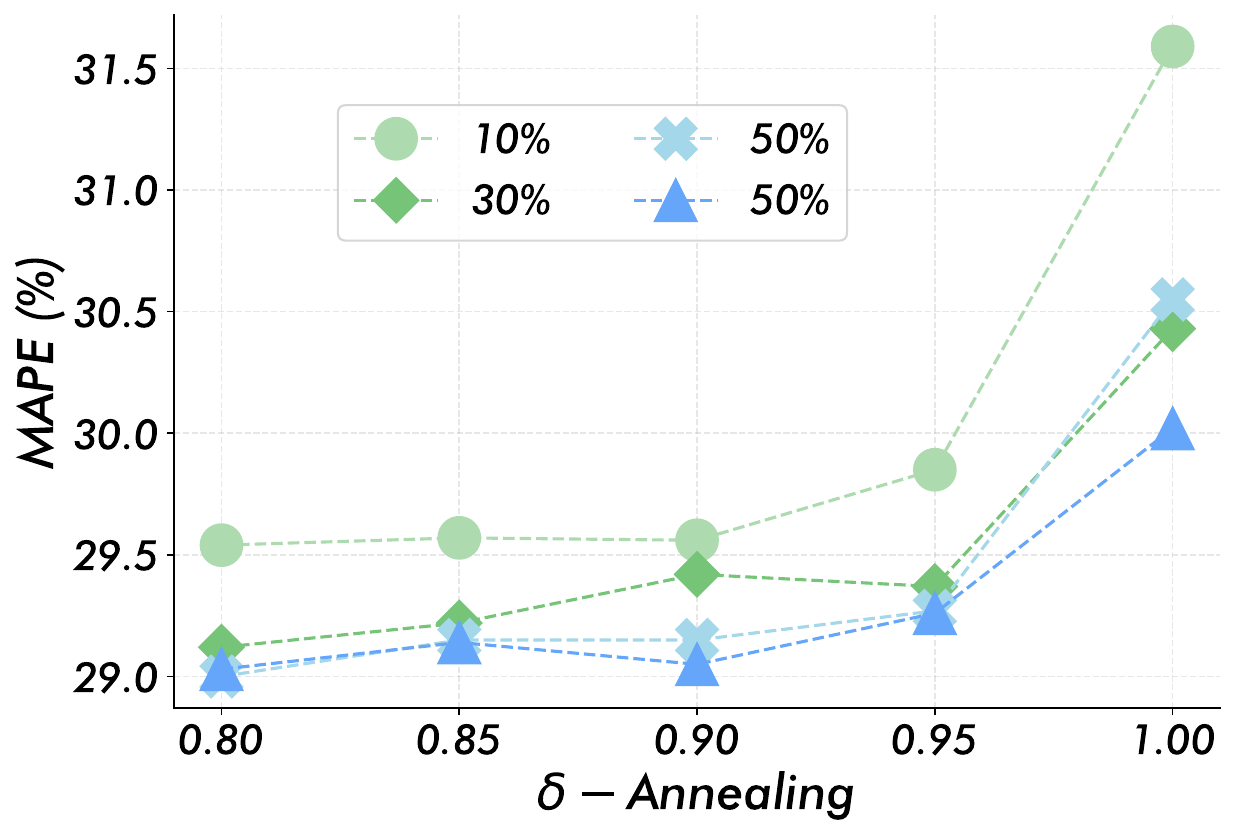}
\end{minipage}
\caption{Ablation and parameter sensitivity analysis on \UrbanEV~dataset under training data retention rates of $\{10\%,30\%,50\%,70\%\}$.}
\label{fig:rq5_mechanism}
\end{figure*}

%% file: tables/rq5_figure_cluster.tex
\begin{figure*}[htbp!]
    \centering
    \includegraphics[width=1.0\linewidth]{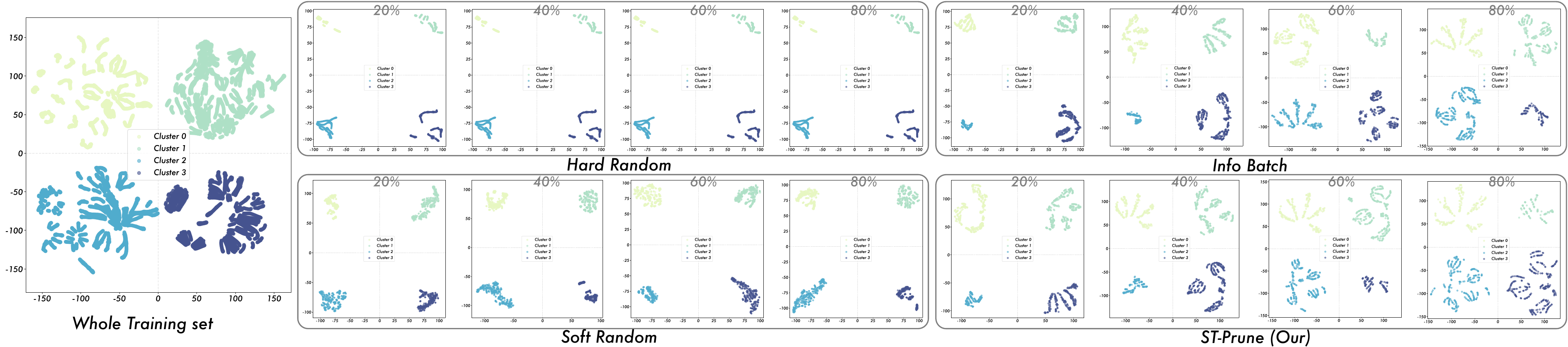}
    \caption{Visualization of t-SNE fixed-corner clustering of the \PemsEight~ distribution with the full training set and a retention rate of 20\%. The right panel shows the evolution of selected subsets through different pruning strategies at different training progresses (20\% to 80\%).}
    \label{fig:rq5_cluster}
\end{figure*}

%% file: sections/06_conclusion.tex
\section{Conclusion and Future Work}

In this paper, we investigate the redundancy inherent in spatio-temporal training datasets and explore effective approaches for efficient dynamic data pruning. We propose \model that uses complexity-informed scoring with stationarity-aware gradient rescaling to prioritize informative samples during training. Extensive experiments confirm its effectiveness. In future work, we aim to explore how to extend this method to continual spatio-temporal  forecasting where the spatial topology evolves over training time.

%% file: appendix/0_appendix_list.tex
\section*{Appendix}
\appendix
\counterwithin{figure}{section}
\counterwithin{table}{section}
\fancypagestyle{appendixfooter}{
  \fancyhf{} 
  \renewcommand{\headrulewidth}{0pt}
  \renewcommand{\footrulewidth}{0pt}
  \fancyfoot[L]{\hyperlink{appendix-start}{{\textit{Go to Appendix Index}}}}
  \fancyfoot[C]{\thepage}
}

\hypertarget{appendix-start}{}
\pagestyle{appendixfooter}


\vspace{1.5em}

\hrule height .8pt
\DoToC
\hrule height .8pt

\vspace{1.5em}

\clearpage

\input{appendix/2_expsetting}

\input{appendix/4_moreresult}

\input{appendix/5_morediscussion}

%% file: appendix/2_expsetting.tex
\section{Experimental Details}

\subsection{Dataset and Protocol Details}\label{appendix_datasets}

\subsubsection{Datasets Details}
Our experiments are carried out on five real-world datasets from diffrent domain. The statistics of these spatio-temporal datasets are shown in Table~\ref{tab:dataset_summary}.

\input{tables/datasets}

We adopt the standard approach widely used in prior works~\citep{li2017diffusion,bai2020adaptive} to determine the graph structure prior. Specifically, we build the adjacency matrix $A$ using a thresholded Gaussian kernel, formulated as:

$$
A_{[ij]} =
\begin{cases}
\exp\!\left(-\frac{d_{ij}^2}{\sigma^2}\right), & \text{if } \exp\!\left(-\frac{d_{ij}^2}{\sigma^2}\right) \geq r \ \text{and}\ i \neq j,\\[6pt]
0, & \text{otherwise},
\end{cases}
$$

where $d_{ij}$ denotes the geographical distance between sensors $i$ and $j$, $\sigma$ is chosen as the standard deviation of all pairwise distances, and $r$ is a predefined similarity threshold used to sparsify the graph. For the specific parameter settings, we follow the paper corresponding to the dataset.

\subsubsection{Protocol Details}

We use different metrics such as MAE, RMSE, MAPE, and CORR. Formally, these metrics are formulated as following:
$$\text{MAE} = \frac{1}{n} \sum_{i=1}^{n} |y_i - \hat{y}_i|,~~~~~~~~~~~~~~~~~~\text{RMSE} = \sqrt{\frac{1}{n} \sum_{i=1}^{n} (y_i - \hat{y}_i)^2},$$
$$\text{MAPE}=\frac{100\%}{n}\sum_{i=1}^{n}{\left|\frac{\hat{y}_i-y_i}{y_i}\right|},~~~~~~~~~~~~~~~~~~\text{Corr} = \frac{\sum_{i=1}^{n} (y_i - \bar{y})(\hat{y}_i - \bar{\hat{y}})}{\sqrt{\sum_{i=1}^{n} (y_i - \bar{y})^2} \sqrt{\sum_{i=1}^{n} (\hat{y}_i - \bar{\hat{y}})^2}},$$
where $n$ represents the indices of all observed samples, $y_i$ denotes the $i$-th actual sample, and $\hat{y_i}$ is the corresponding prediction.

\input{tables/data_sta_vis}

\subsection{Baseline and Parameter Details}\label{appendix_baseline}

\subsubsection{Baseline Details}

In this subsection, we describe in detail the advanced dataset pruning and selection methods and different spatiotemporal backbone models that we use in our default evaluation.

\textbf{Data Pruning and Selection Methods.} To ensure fair use of these baselines, we use a popular coreset selection benchmark libraries: \url{https://github.com/PatrickZH/DeepCore} and adapt the respective methods to ST forecasting tasks.

\begin{itemize}[leftmargin=*,itemsep=1.5em,parsep=1.5pt]
    \item \textit{{Static Dataset Pruning}}:
    \begin{itemize}[leftmargin=*,parsep=2pt]
        \item[\dag] \textit{{Random-based}}: 
        \begin{itemize}[leftmargin=*,itemsep=1em,parsep=1pt]
            \item[\ddag] \textbf{{Hard Random}}: Random selection before training.
        \end{itemize}
        \item[\dag] \textit{{Geometry-based}}: 
        \begin{itemize}[leftmargin=*,itemsep=0.5em,parsep=1pt]
            \item[\ddag] \textbf{{\CD}~\citep{agarwal2015construction}}: Adapted the diversity metric from image feature space to spatio-temporal feature space, using spatio-temporal embedding vectors to calculate inter-sample distances for selecting representative patterns.
            \item[\ddag] \textbf{{\Herding}~\citep{welling2009herding}}: Modified prototype selection from classification labels to regression targets, finding core samples in spatio-temporal feature space that represent different trends.
            \item[\ddag] \textbf{{\Kmeans}~\citep{sener2018active}}: Adapted \Kmeans~clustering from image features to spatio-temporal sequence features, clustering based on temporal patterns and spatial correlations in samples for center selection.
        \end{itemize}
        \item[\dag] \textit{{Uncertainty-based}}: 
        \begin{itemize}[leftmargin=*,itemsep=0.5em,parsep=1pt]
            \item[\ddag] \textbf{{\LeastConfidence}~\citep{coleman2019selection}}:Replaced classification probability confidence with regression prediction uncertainty estimation, selecting spatio-temporal samples with maximum prediction variance.
            \item[\ddag] \textbf{{\Entropy}~\citep{coleman2019selection}}: Adapted entropy calculation from classification probability distributions to regression prediction distributions, selecting spatio-temporal sequence samples with maximum uncertainty-based information content.
            \item[\ddag] \textbf{{\Margin}~\citep{coleman2019selection}}: Changed from classification decision boundary distances to regression prediction interval widths, selecting spatio-temporal samples with maximum prediction confidence intervals.
        \end{itemize}
        \item[\dag] \textit{{Loss-based}}: 
        \begin{itemize}[leftmargin=*,itemsep=0.5em,parsep=1pt]
            \item[\ddag] \textbf{{\Forgetting}~\citep{toneva2018empirical}}: Replaced classification accuracy forgetting statistics with regression error change statistics, tracking fluctuation patterns of spatio-temporal sequence samples' prediction errors during training.
        \end{itemize}
        \item[\dag] \textit{{Decision boundary-based}}: 
        \begin{itemize}[leftmargin=*,itemsep=0.5em,parsep=1pt]
            \item[\ddag] \textbf{{\GraNd}~\citep{paul2021deep}}: Adapted gradient norm calculation from classification loss to regression loss, selecting samples with greatest impact on the model based on spatio-temporal prediction model parameter gradient magnitudes.
        \end{itemize}
        \item[\dag] \textit{{Bi-level optimization-based}}: 
        \begin{itemize}[leftmargin=*,itemsep=0.5em,parsep=1pt]
            \item[\ddag] \textbf{{\Glister}~\citep{killamsetty2021glister}}: Replaced classification loss with regression loss in the bi-level optimization framework, selecting most valuable training samples through gradient matching for spatio-temporal prediction tasks.
        \end{itemize}
        \item[\dag] \textit{{Submodular function-based}}: 
        \begin{itemize}[leftmargin=*,itemsep=0.5em,parsep=1pt]
            \item[\ddag] \textbf{{\GraphCut}~\citep{iyer2021submodular}}: Replaced similarity-based construction with spatio-temporal sequence similarity-based construction, performing partitioning based on spatio-temporal correlations to select representative samples.
            \item[\ddag] \textbf{{\FaLo}~\citep{iyer2021submodular}}: Adapted the facility location problem from classification to regression tasks, finding facility points in spatio-temporal feature space that cover different patterns as core samples.
        \end{itemize}
    \end{itemize}
    \item \textit{{Dynamic Dataset Pruning}}:
    \begin{itemize}[leftmargin=*,parsep=2pt]
        \item[\dag] \textit{{Random-based}}: 
        \begin{itemize}[leftmargin=*,itemsep=0.5em,parsep=1pt]
            \item[\ddag] \textbf{{\SoftRandom}}: Random selection during training.
        \end{itemize}
        \item[\dag] \textit{{Uncertainty-based}}: 
        \begin{itemize}[leftmargin=*,itemsep=0.5em,parsep=1pt]
            \item[\ddag] \textbf{{\ThetaGreedy}~\citep{raju2021accelerating}}: Adapted greedy selection strategy from classification models to regression model parameter changes, dynamically selecting samples most beneficial for spatio-temporal prediction model training.
            \item[\ddag] \textbf{{\UCB}~\citep{raju2021accelerating}}: Adapted multi-armed bandit classification reward mechanism to regression task rewards, dynamically balancing exploration and exploitation in sample selection based on prediction error reduction.
        \end{itemize}
        \item[\dag] \textit{{Loss-based}}: 
        \begin{itemize}[leftmargin=*,itemsep=0.5em,parsep=1pt]
            \item[\ddag] \textbf{{\InfoBatch}~\citep{qin2023infobatch}}: Adapted dynamic batch selection from classification loss-based to spatiotemporal regression loss-based selection, dynamically adjusting training batch composition according to the information content of spatio-temporal sequence samples.
        \end{itemize}
    \end{itemize}
\end{itemize}

\textbf{Spatio-Temporal Neural Network.} To ensure fair use of these models, we use an implementation of the popular spatio-temporal neural network benchmark library: \url{https://github.com/liuxu77/LargeST}, adapted to the above data pruning and selection repository.

\begin{itemize}[leftmargin=*,parsep=5pt]
    \item \textit{{Graph-based}}: 
    \begin{itemize}[leftmargin=*,itemsep=1em,parsep=1pt]
        \item[\dag] \textbf{{\GWNet}~\citep{wu2019graph}}: \GWNet~is a graph-based deep learning framework tailored for spatio-temporal forecasting. It leverages graph convolutional networks to model complex spatial correlations among nodes and employs dilated causal convolutions to capture long-range temporal dependencies, enabling effective representation of spatial-temporal dynamics.
    \end{itemize}
    \item \textit{{MLP-based}}: 
    \begin{itemize}[leftmargin=*,itemsep=1em,parsep=1pt]
        \item[\dag] \textbf{{\STID}~\citep{shao2022spatial}}: \STID~serves as a lightweight MLP-based baseline for spatio-temporal prediction. By explicitly incorporating spatial and temporal identity embeddings, it alleviates the indistinguishability problem in spatio-temporal data. Its parameter-efficient MLP architecture achieves competitive forecasting accuracy with low computational overhead.
    \end{itemize}
    \item \textit{{Transformer-based}}: 
    \begin{itemize}[leftmargin=*,itemsep=1em,parsep=1pt]
        \item[\dag] \textbf{{\STAEformer}~\citep{liu2023spatio}}: \STAEformer~is a effective transformer-based model that adapts the vanilla transformer architecture to the spatio-temporal domain. It introduces adaptive spatial-temporal embeddings and attention mechanisms to dynamically capture spatial dependencies and temporal evolution, pushing transformer models to high-competitive performance for spatio-temporal forecasting tasks.
    \end{itemize}
\end{itemize}

\textbf{Spatio-Temporal Foundation Model.} We also applied our method to the recently popular spatio-temporal foundation models to evaluate scalability.
We followed and used the open-source code repository and pre-trained datasets of OpenCity~\citep{Li2025OpenSF} from: \url{https://github.com/HKUDS/OpenCity}.

\begin{itemize}[leftmargin=*,itemsep=1em,parsep=1pt]
    \item[\dag] \textbf{{\Opencity}~\citep{Li2025OpenSF}}: 
    \Opencity~is a large-scale foundation model designed for general-purpose spatio-temporal learning in urban environments. It is pre-trained on massive, heterogeneous urban data spanning multiple cities and domains, enabling strong cross-city and cross-task generalization. 
\end{itemize}

\subsubsection{Parameter Details}

Detailed hyperparameters settings are shown in Table~\ref{tab:Hyperparameters}. We use the same parameter configurations for our \model, along with the other baseline methods according to the recommendation of previous papers~\citep{guo2022deepcore,liu2024largest}. All experiments are conducted on a Linux server equipped with a 1 × Intel(R) Xeon(R) Gold 6248R CPU @ 3.00GHz (512GB memory) and 8 × NVIDIA A100 (80GB memory) GPUs. To carry out benchmark testing experiments, all baselines and our method are set to run 100 epochs by default without early stopping mechanism to ensure fairness of efficiency experiments. 

\input{tables/parameter}

%% file: tables/datasets.tex
\begin{table}[htbp!]
\centering
\tabcolsep=0.65mm
\renewcommand{\arraystretch}{1.4}
  \caption{Summary of datasets used for our experiments. Degree: the average degree of each node. Data Points: multiplication of nodes and frames. \textcolor{blue}{M}: million ($10^6$).}
  \vspace{2mm}
  \begin{tabular}{cc|ccccc}
    \toprule
     Source & Dataset & Nodes & Time Range & Frames & Sampling Rate & Data Points \\
    \hline
    \multirow{1}{*}{~\citep{song2020spatial}} 
     & PEMS08 & 170 & 07/01/2016 -- 08/31/2016 & 17,856 & 5 minutes & 3.04\textcolor{blue}{M} \\
    \multirow{1}{*}{~\citep{li2025urbanev}} 
     & UrbanEV  & 275 & 09/01/2022 -- 02/28/2023 & 4344 & 1 hour & 1.19\textcolor{blue}{M} \\
     \hline
    \multirow{3}{*}{LargeST~\citep{liu2024largest}}
    & GLA & 3,834 & 01/01/2019 -- 12/31/2019 & 35,040 & 15 minutes & 134.3\textcolor{blue}{M} \\
    & GBA & 2,352 & 01/01/2019 -- 12/31/2019 & 35,040 & 15 minutes & 88.7\textcolor{blue}{M} \\
    & SD & 716 & 01/01/2019 -- 12/31/2019 & 35,040 & 15 minutes & 25.1\textcolor{blue}{M} \\
    \bottomrule
  \end{tabular}
  \label{tab:dataset_summary}
\end{table}

%% file: tables/data_sta_vis.tex
\begin{figure*}[htbp!]
\centering
\begin{minipage}[t]{0.3\linewidth}
  \centering
  \includegraphics[width=\linewidth]{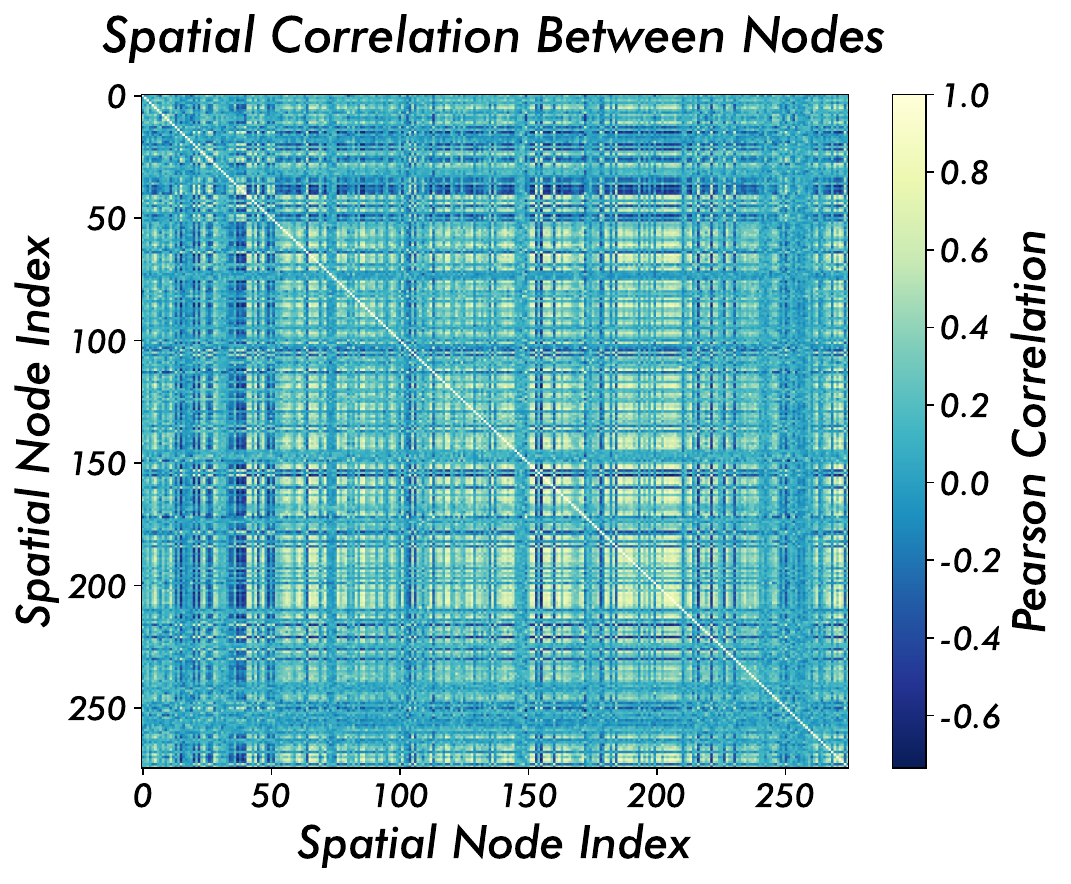}
\end{minipage}\hfill
\begin{minipage}[t]{0.3\linewidth}
  \centering
  \includegraphics[width=\linewidth]{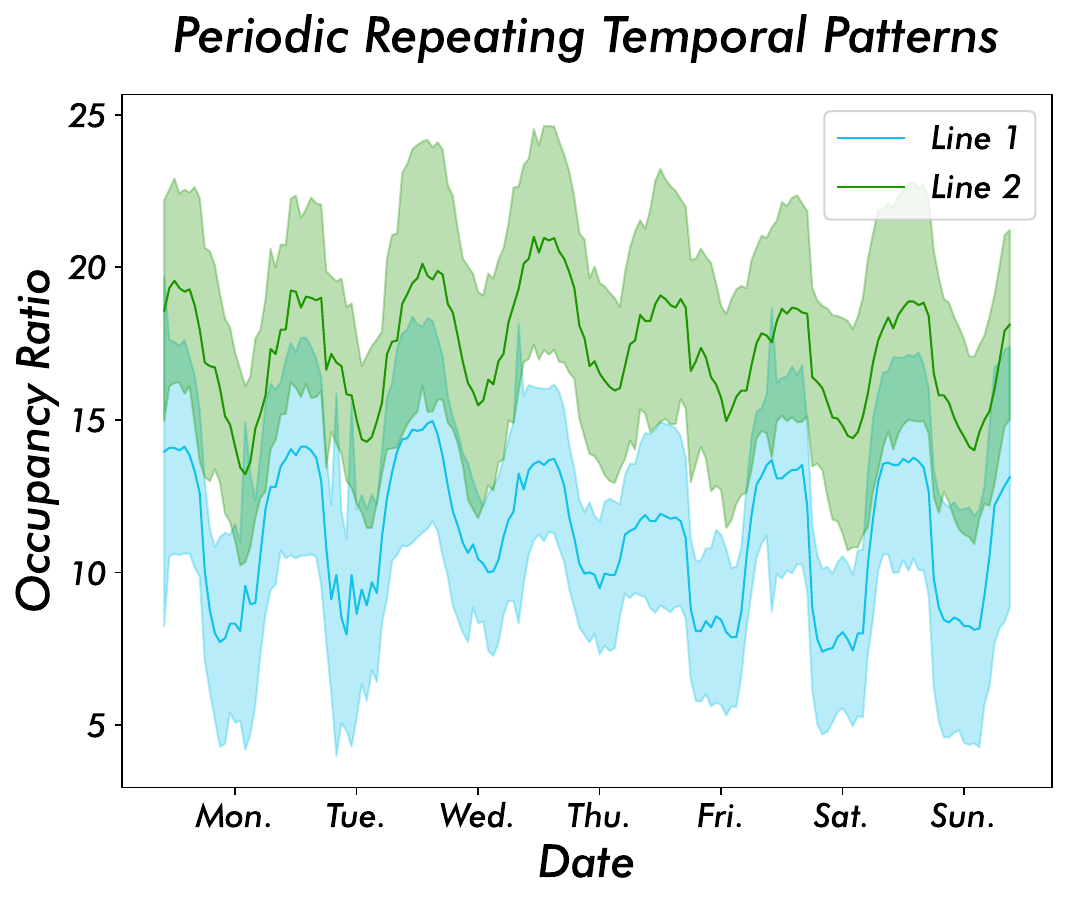}
\end{minipage}\hfill
\begin{minipage}[t]{0.3\linewidth}
  \centering
  \includegraphics[width=\linewidth]{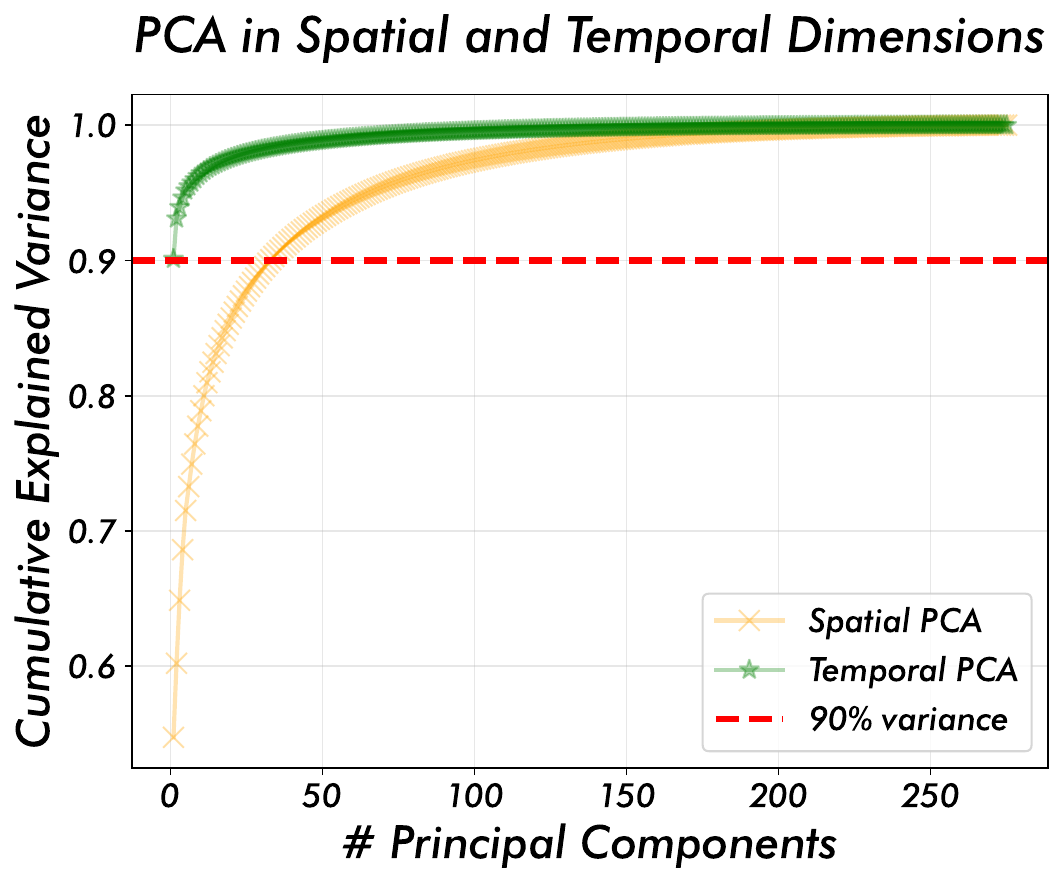}
\end{minipage}
\vspace{-2mm}
\caption{The spatio-temporal redundancy characteristics and statistical properties of \textsc{UrbanEV}~\citep{li2025urbanev}.}
\label{fig:data_sta_vis}
\end{figure*}

\begin{figure*}[htbp!]
\centering
\begin{minipage}[t]{0.3\linewidth}
  \centering
  \includegraphics[width=\linewidth]{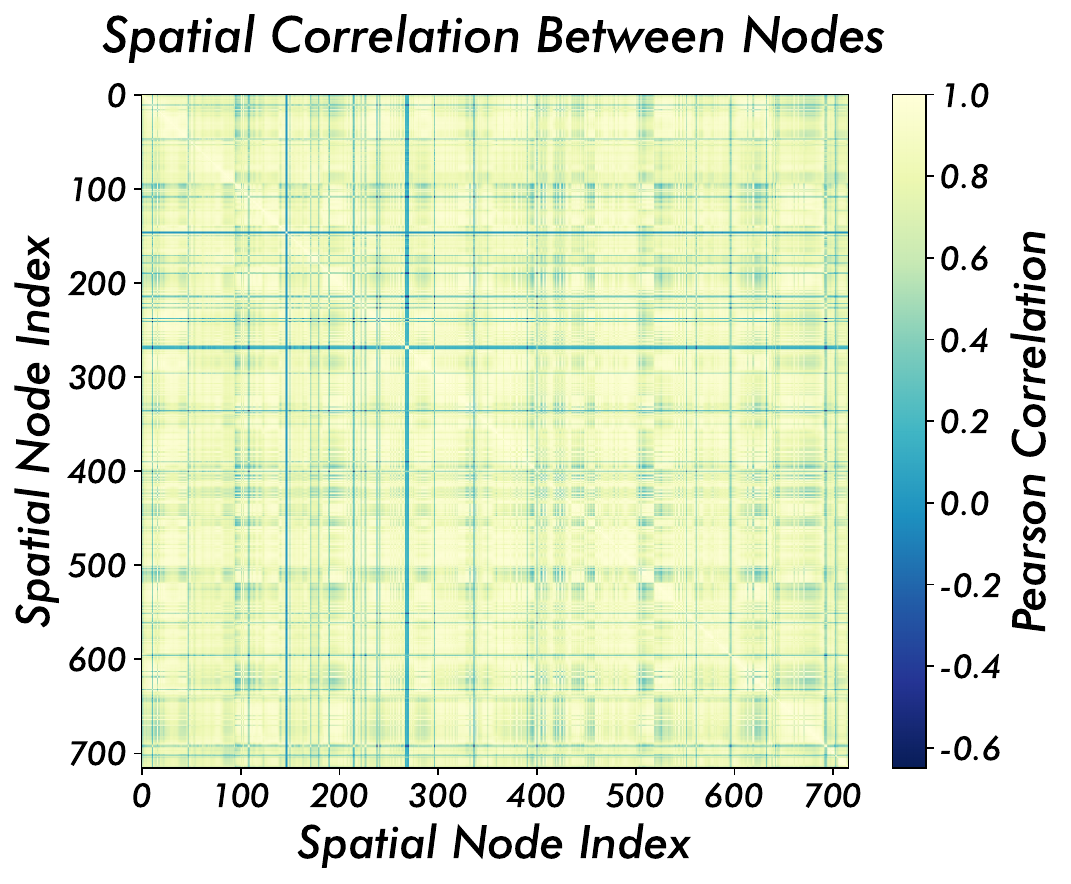}
\end{minipage}\hfill
\begin{minipage}[t]{0.3\linewidth}
  \centering
  \includegraphics[width=\linewidth]{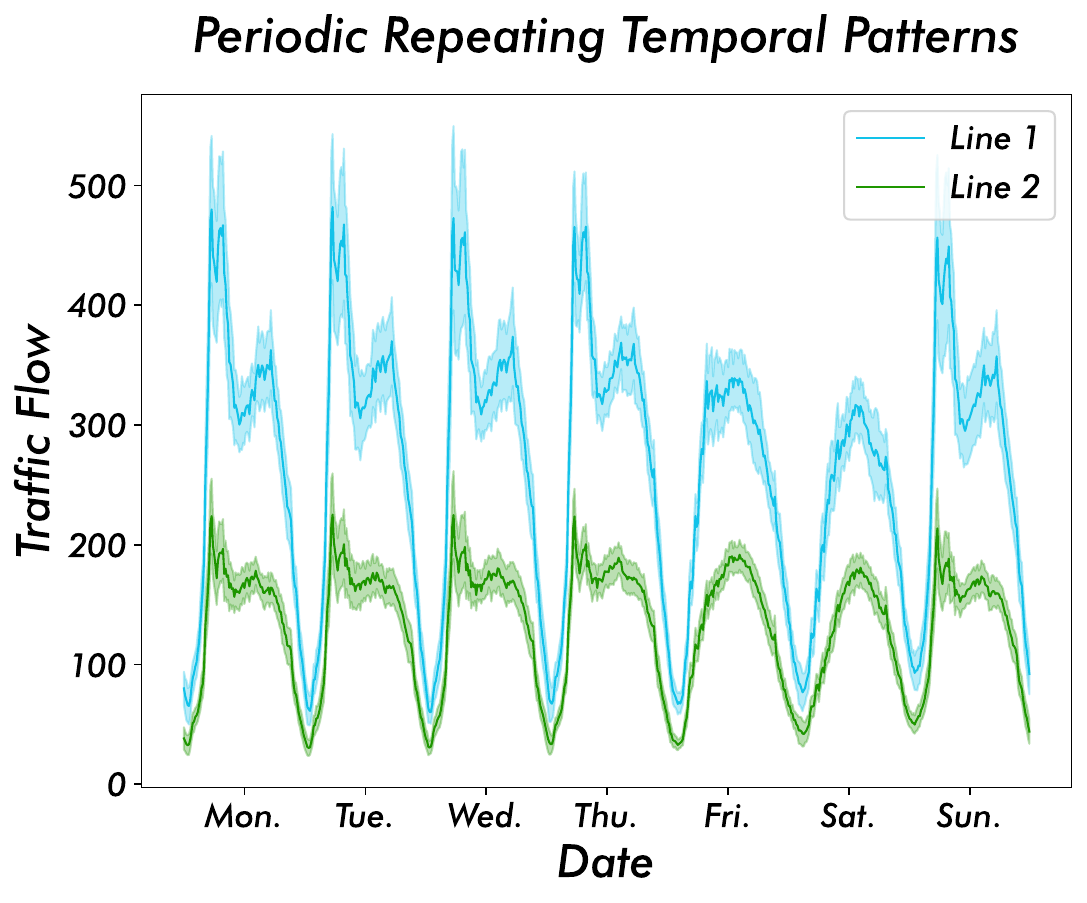}
\end{minipage}\hfill
\begin{minipage}[t]{0.3\linewidth}
  \centering
  \includegraphics[width=\linewidth]{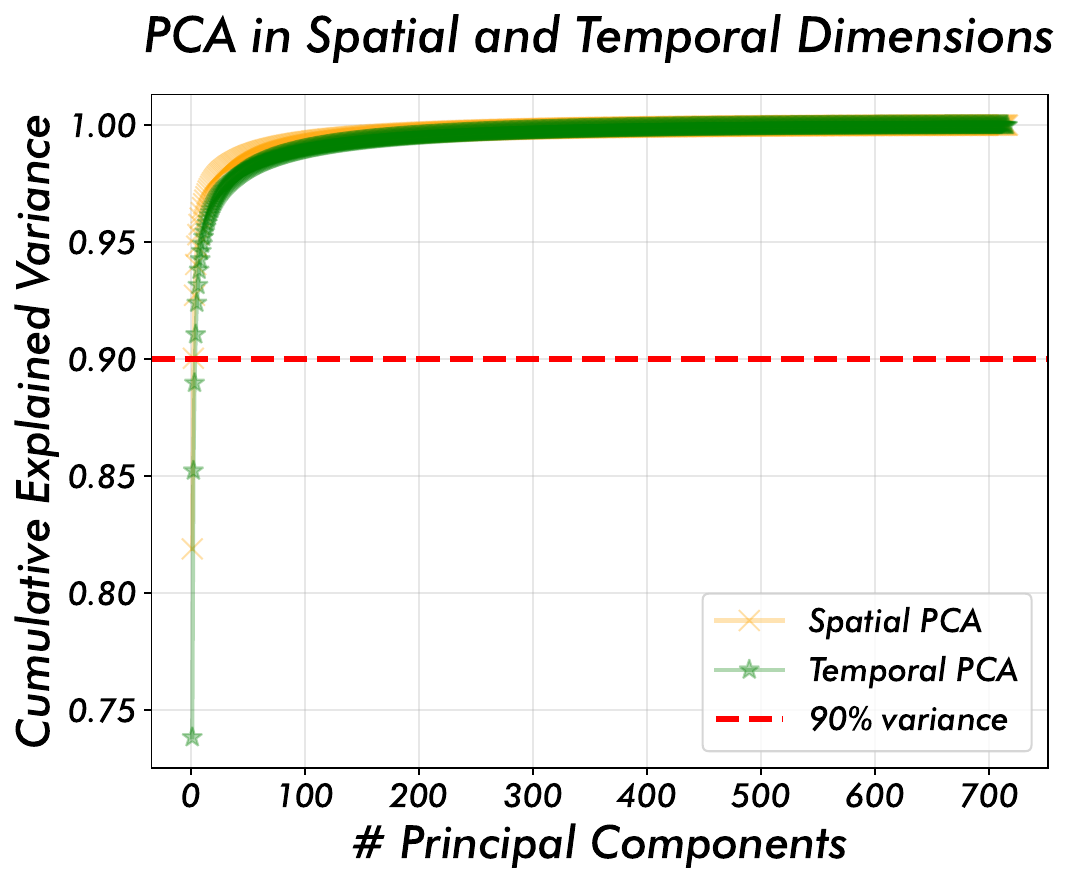}
\end{minipage}
\vspace{-2mm}
\caption{The spatio-temporal redundancy characteristics and statistical properties of \textsc{SD}~\citep{liu2024largest}.}
\label{fig:data_sta_vis}
\end{figure*}

\begin{figure*}[htbp!]
\centering
\begin{minipage}[t]{0.3\linewidth}
  \centering
  \includegraphics[width=\linewidth]{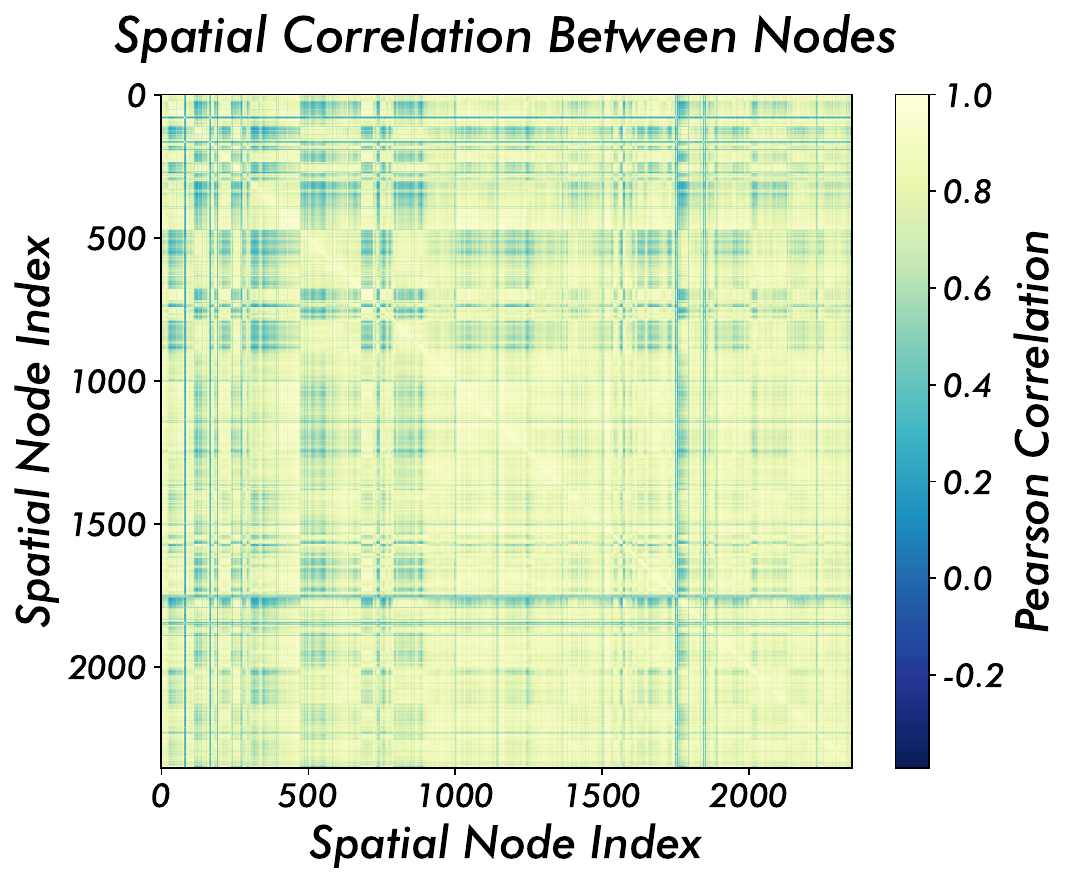}
\end{minipage}\hfill
\begin{minipage}[t]{0.3\linewidth}
  \centering
  \includegraphics[width=\linewidth]{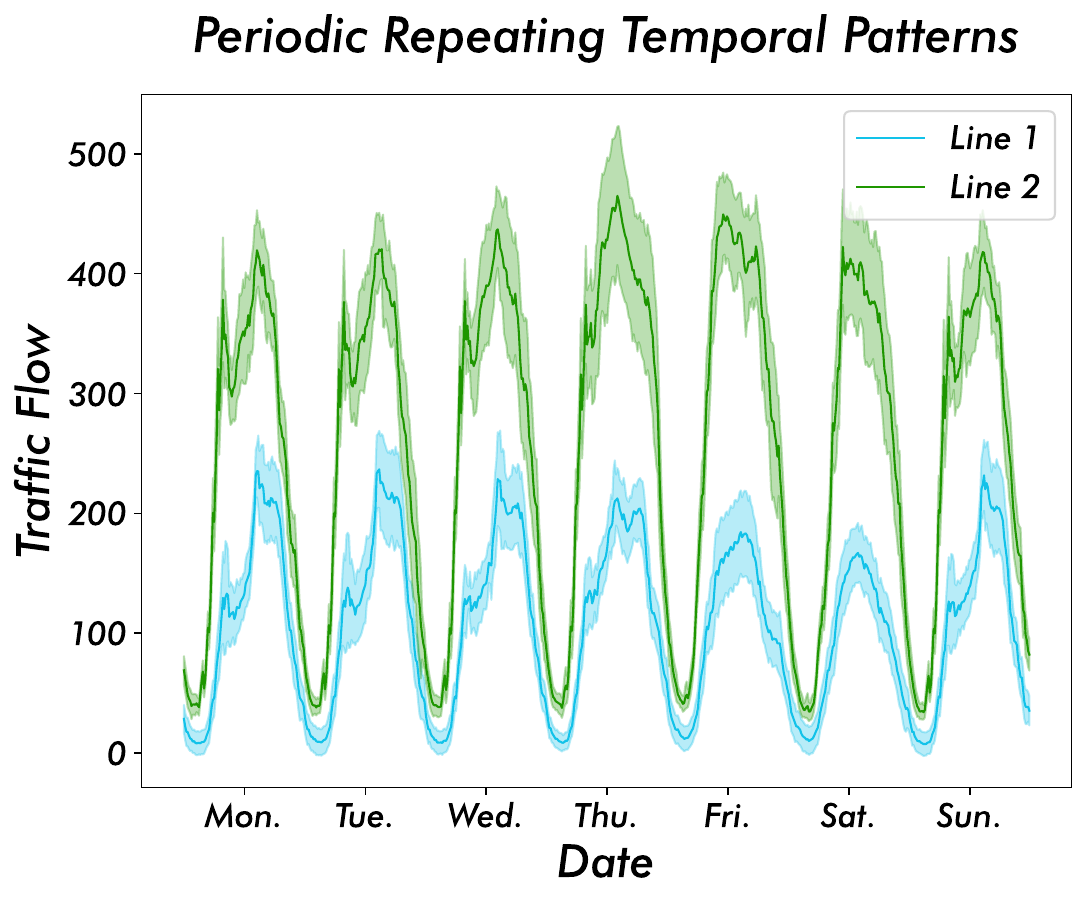}
\end{minipage}\hfill
\begin{minipage}[t]{0.3\linewidth}
  \centering
  \includegraphics[width=\linewidth]{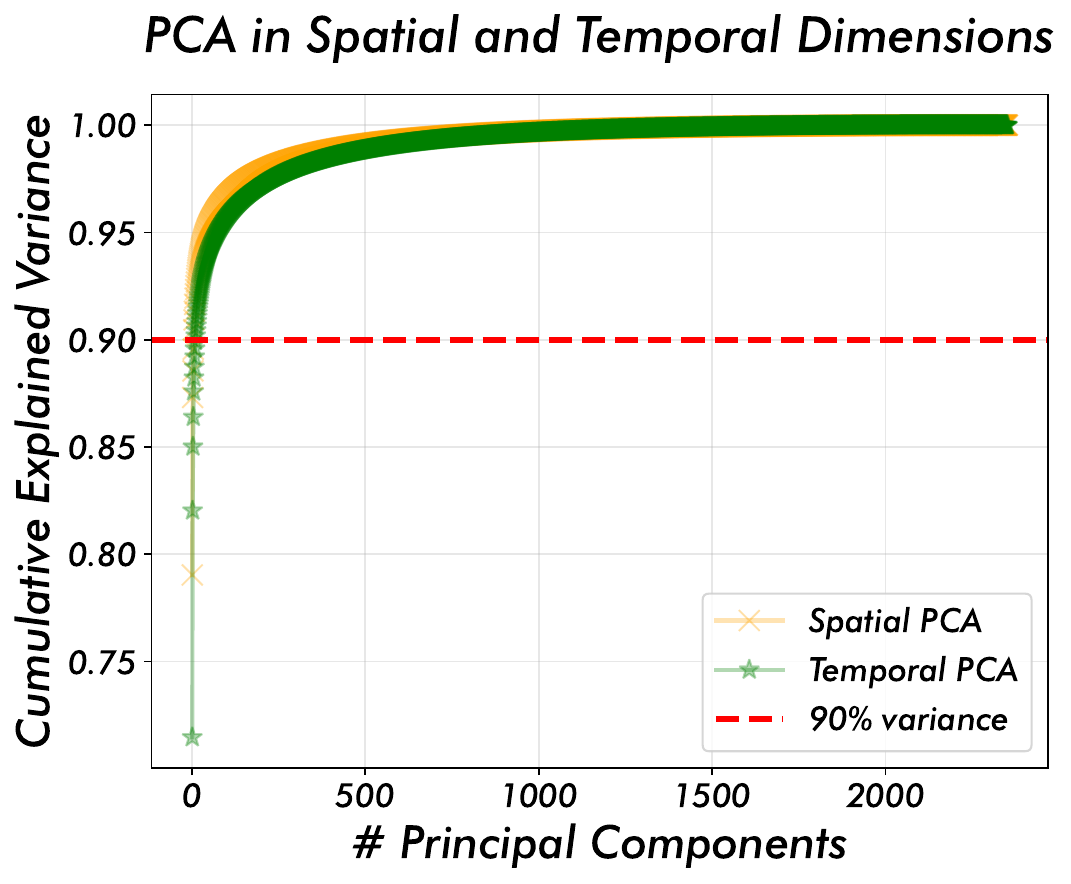}
\end{minipage}
\vspace{-2mm}
\caption{The spatio-temporal redundancy characteristics and statistical properties of \textsc{GBA}~\citep{liu2024largest}.}
\label{fig:data_sta_vis}
\end{figure*}

\begin{figure*}[htbp!]
\centering
\begin{minipage}[t]{0.3\linewidth}
  \centering
  \includegraphics[width=\linewidth]{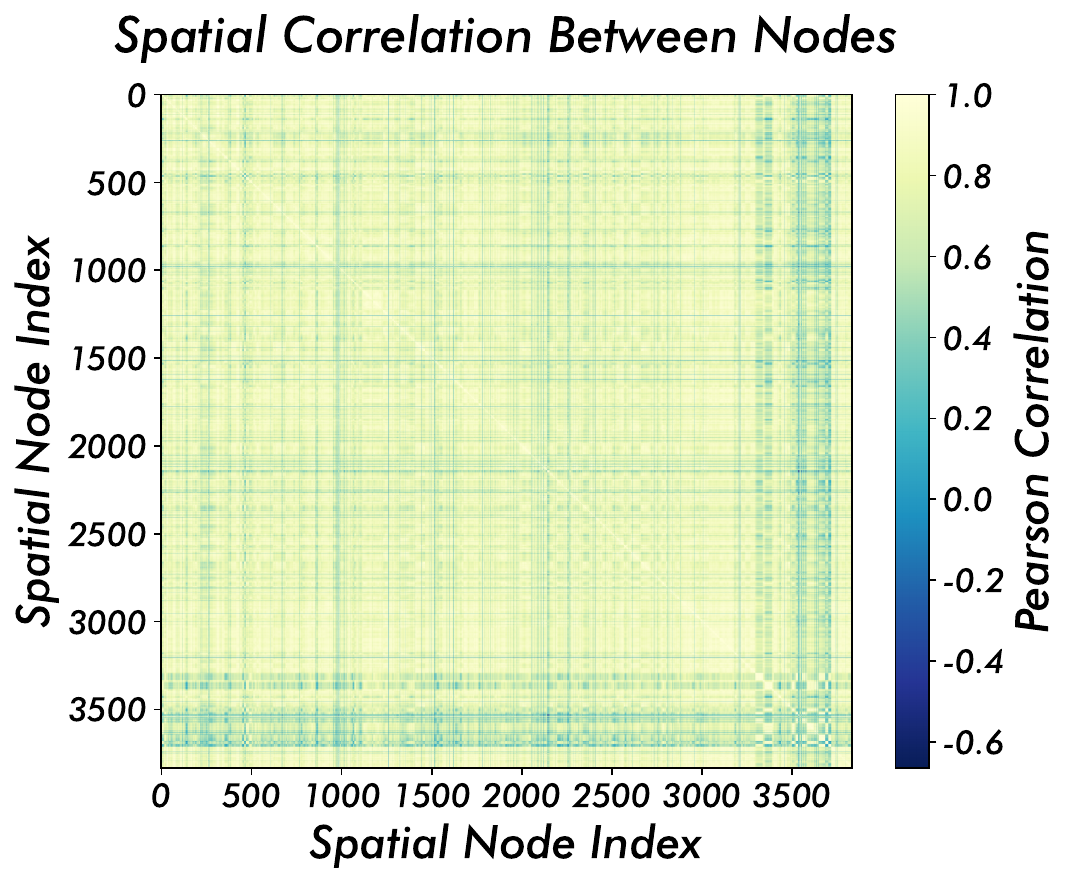}
\end{minipage}\hfill
\begin{minipage}[t]{0.3\linewidth}
  \centering
  \includegraphics[width=\linewidth]{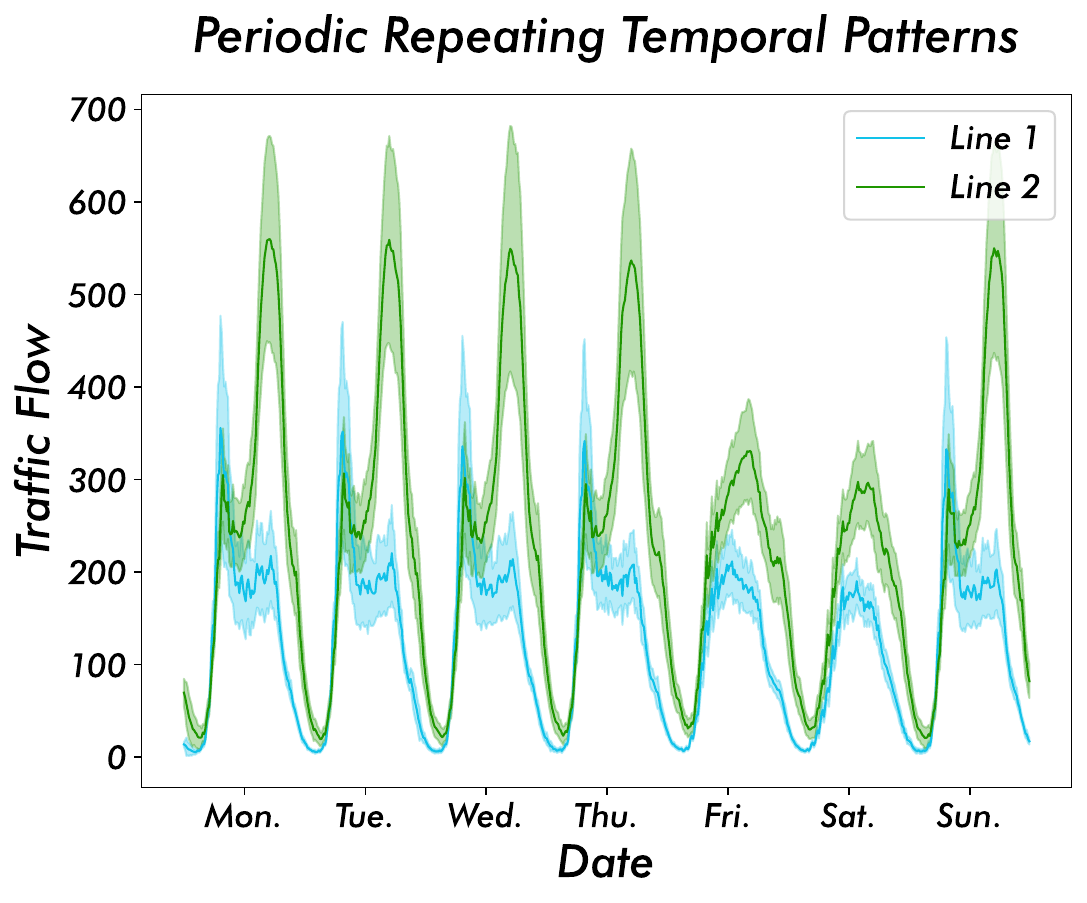}
\end{minipage}\hfill
\begin{minipage}[t]{0.3\linewidth}
  \centering
  \includegraphics[width=\linewidth]{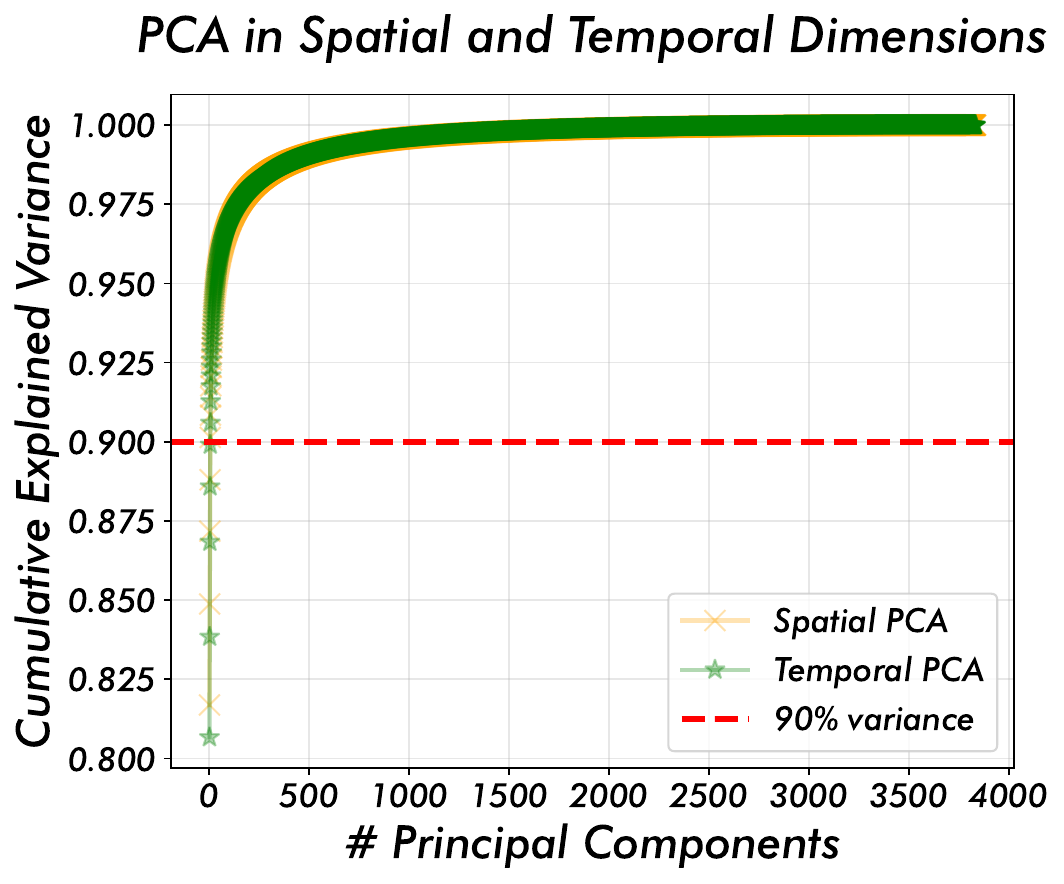}
\end{minipage}
\vspace{-2mm}
\caption{The spatio-temporal redundancy characteristics and statistical properties of \textsc{GLA}~\citep{liu2024largest}.}
\label{fig:data_sta_vis}
\end{figure*}

%% file: tables/parameter.tex
\begin{table}[htbp!]
\centering
\vspace{-8mm}
\caption{Hyperparameters setting.}
\label{tab:Hyperparameters}
\fontsize{9}{12} \selectfont
\setlength{\tabcolsep}{3.0mm}{}	
\begin{tabular}{c|c}
     \toprule
     Training epochs & 100\\
     Selection epochs & 10\\
     Batch size & 256\\
     Data Fraction & 0.1 / 0.3 / 0.5 / 0.7\\
     \midrule
     Optimizer & SGD\\
     Learning rate & 1e-3\\
     Minimum Learning rate & 1e-4\\
     Momentum & 0.9\\
     Weight Decay & 1e-4\\
     \midrule
     Scheduler & CosineAnnealingLR\\
     Gamma & 0.5\\
     Step Size & 50\\
     \midrule
     Loss Function & MAE\\
      \midrule
      $\lambda$ & 0.5 \\
     $\delta$ & 0.9 \\
     \bottomrule
\end{tabular}
\vspace{-6mm}
\end{table}

%% file: appendix/4_moreresult.tex
\section{More Results}\label{appendix_moreresult}

We also provide MAE and RMSE comparisons of the performance of our method \model with the state-of-the-art dataset pruning and selection methods when $\{10\%, 30\%, 50\%, 70\%\}$ of the full set remains, as shown in Tables~\ref{tab:rq1_mae} and~\ref{tab:rq1_rmse}.

\input{tables/appendix_rq1_mae}

\input{tables/appendix_rq1_rmse}

%% file: tables/appendix_rq1_mae.tex
\begin{table*}[htbp]
    \caption{Performance comparison to state-of-the-art dataset pruning methods when remaining $\{10\%,30\%,50\%,70\%\}$ of the full set. All methods are trained using \textbf{GWNet}, and the reported metric MAE represent the average of \textbf{five runs}. }
\label{tab:rq1_mae}
    \centering
    \footnotesize
    \renewcommand{\arraystretch}{1.2}
    \setlength{\tabcolsep}{3pt}
    \resizebox{\textwidth}{!}{
    \begin{tabular}{cc|cccc|cccc}
    \toprule
    \multirow{3}{*}{} & Dataset  & \multicolumn{4}{c|}{\textsc{Pems08} (MAE $\downarrow$)} & \multicolumn{4}{c}{\textsc{UrbanEV} (MAE $\downarrow$)}  \\
    \midrule
    & Remaining Ratio \% & 10 & 30 & 50 & 70  & 10 & 30 & 50 & 70 \\ \midrule
    \parbox[t]{4mm}{\multirow{13}{*}{\rotatebox[origin=c]{90}{Static}}}
    
    & Hard Random
    & 19.92\blue{16.90\%} & 18.75\blue{10.04\%} & 18.24\blue{7.04\%} & 17.92\blue{5.16\%}
    & 4.23\blue{18.82\%} & 3.96\blue{11.24\%} & 3.82\blue{7.30\%} & 3.76\blue{5.62\%}
    \\
    
    & CD~\citep{agarwal2020contextual}
    & 19.99\blue{17.31\%} & 18.50\blue{8.57\%} & 18.12\blue{6.34\%} & 17.86\blue{4.81\%}
    & 4.34\blue{21.91\%} & 3.95\blue{10.96\%} & 3.81\blue{7.02\%} & 3.79\blue{6.46\%}
    \\
    
    & Herding~\citep{welling2009herding}
    & 20.14\blue{18.19\%} & 18.51\blue{8.63\%} & 18.16\blue{6.57\%} & 17.84\blue{4.69\%}
    & 4.37\blue{22.75\%} & 3.92\blue{10.11\%} & 3.88\blue{8.99\%} & 3.74\blue{5.06\%}
    \\
    
    & K-Means~\citep{sener2018active}
    & 20.21\blue{18.60\%} & 18.59\blue{9.10\%} & 18.28\blue{7.28\%} & 17.81\blue{4.52\%}
    & 4.33\blue{21.63\%} & 4.00\blue{12.36\%} & 3.81\blue{7.02\%} & 3.77\blue{5.90\%}
    \\
    
    & Least Confidence~\citep{coleman2019selection}
    & 20.61\blue{20.95\%} & 19.48\blue{14.32\%} & 18.14\blue{6.46\%} & 18.11\blue{6.28\%}
    & 4.22\blue{18.54\%} & 4.47\blue{25.56\%} & 3.82\blue{7.30\%} & 3.71\blue{4.21\%}
    \\
    
    & Entropy~\citep{coleman2019selection}
    & 20.65\blue{21.19\%} & 18.49\blue{8.51\%} & 18.10\blue{6.22\%} & 17.88\blue{4.93\%} 
    & 4.32\blue{21.35\%} & 3.96\blue{11.24\%} & 3.81\blue{7.02\%} & 3.76\blue{5.62\%}
    \\
    
    & Margin~\citep{coleman2019selection}
    & 20.42\blue{19.84\%} & 18.62\blue{9.27\%} & 18.13\blue{6.40\%} & 17.97\blue{5.46\%}
    & 4.25\blue{19.38\%} & 4.02\blue{12.92\%} & 4.00\blue{12.36\%} & 3.72\blue{4.49\%}
    \\
    
    & Forgetting~\citep{toneva2018empirical}
    & 19.77\blue{16.02\%} & 18.84\blue{10.56\%} & 18.16\blue{6.57\%} & 17.81\blue{4.52\%}
    & 4.24\blue{19.10\%} & 3.97\blue{11.52\%} & 3.80\blue{6.74\%} & 3.74\blue{5.06\%}
    \\
    
    & GraNd~\citep{paul2021deep}
    & 20.07\blue{17.78\%} & 18.48\blue{8.45\%} & 18.30\blue{7.39\%} & 17.85\blue{4.75\%}
    & 4.24\blue{19.10\%} & 3.99\blue{12.08\%} & 3.85\blue{8.15\%} & 3.78\blue{6.18\%}
    \\
    
    & Cal~\citep{margatina2021active}
    & 19.71\blue{15.67\%} & 18.66\blue{9.51\%} & 18.77\blue{10.15\%} & 18.14\blue{6.46\%}
    & 4.22\blue{18.54\%} & 3.99\blue{12.08\%} & 3.83\blue{7.58\%} & 3.75\blue{5.34\%}
    \\
    
    & Glister~\citep{killamsetty2021glister}
    & 20.71\blue{21.54\%} & 19.10\blue{12.09\%} & 18.87\blue{10.74\%} & 18.40\blue{7.98\%}
    & 4.29\blue{20.51\%} & 4.13\blue{16.01\%} & 3.90\blue{9.55\%} & 3.78\blue{6.18\%}
    \\
    
    & GraphCut~\citep{iyer2021submodular}
    & 20.21\blue{18.60\%} & 18.55\blue{8.86\%} & 18.01\blue{5.69\%} & 17.78\blue{4.34\%}
    & 4.29\blue{20.51\%} & 3.95\blue{10.96\%} & 3.84\blue{7.87\%} & 3.75\blue{5.34\%}
    \\
    
    & FaLo~\citep{iyer2021submodular}
    & 19.97\blue{17.19\%} & 18.57\blue{8.98\%} & 18.19\blue{6.75\%} & 17.95\blue{5.34\%}
    & 4.36\blue{22.47\%} & 3.96\blue{11.24\%} & 3.85\blue{8.15\%} & 3.79\blue{6.46\%}
    \\
    
    \midrule
    \parbox[t]{4mm}{\multirow{5}{*}{\rotatebox[origin=c]{90}{Dynamic}}}
    & Soft Random
    & 21.57\blue{26.58\%} & 18.86\blue{10.68\%} & 18.49\blue{8.51\%} & 17.57\blue{3.11\%}
    & 4.25\blue{19.38\%} & 3.86\blue{8.43\%} & 3.74\blue{5.06\%} & 3.69\blue{3.65\%}
    \\

    & $\epsilon$-greedy~\citep{raju2021accelerating}
    & 20.25\blue{18.84\%} & 18.82\blue{10.45\%} & 18.43\blue{8.16\%} & 17.97\blue{5.46\%}
    & 4.15\blue{16.57\%} & 3.84\blue{7.87\%} & 3.85\blue{8.15\%} & 3.77\blue{5.90\%}
    \\

    & UCB~\citep{raju2021accelerating}
    & 20.24\blue{18.78\%} & 18.95\blue{11.21\%} & 18.34\blue{7.63\%} & 18.05\blue{5.93\%}
    & 4.14\blue{16.29\%} & 3.83\blue{7.58\%} & 3.82\blue{7.30\%} & 3.71\blue{4.21\%}
    \\

    & InfoBatch~\citep{qin2023infobatch}
    & 19.03\blue{11.68\%} & 18.55\blue{8.86\%} & 18.07\blue{6.04\%} & 17.92\blue{5.16\%} 
    & 4.03\blue{13.20\%} & 3.99\blue{12.08\%} & 3.67\blue{3.08\%} & 3.66\blue{2.81\%} 
    \\

    & \model (\textcolor{orange}{Our})
    & \colorbox[HTML]{DAE8FC}{\textbf{18.32\blue{\textbf{7.51}}}} 
    & \colorbox[HTML]{DAE8FC}{\textbf{17.91\blue{\textbf{5.11}}}} 
    & \colorbox[HTML]{DAE8FC}{\textbf{17.63\blue{\textbf{3.46}}}} 
    & \colorbox[HTML]{DAE8FC}{\textbf{17.63\blue{\textbf{3.46}}}}

    & \colorbox[HTML]{DAE8FC}{\textbf{3.63\blue{\textbf{1.97}}}} 
    &  \colorbox[HTML]{DAE8FC}{\textbf{3.62\blue{\textbf{1.69}}}} 
    &   \colorbox[HTML]{DAE8FC}{\textbf{3.58\blue{\textbf{0.56}}}} 
    & \colorbox[HTML]{DAE8FC}{\textbf{3.56\red{\textbf{0.00}}}}

    \\

    \midrule
    \multicolumn{2}{c|}{Whole Dataset} & \multicolumn{4}{c|}{17.04$_{\pm0.23}$} & \multicolumn{4}{c}{3.56$_{\pm0.09}$} \\
    \bottomrule
    \end{tabular}}
    \vspace{-1.5em}
\end{table*}

%% file: tables/appendix_rq1_rmse.tex
\begin{table*}[htbp!]
    \caption{Performance comparison to state-of-the-art dataset pruning methods when remaining $\{10\%,30\%,50\%,70\%\}$ of the full set. All methods are trained using \textbf{GWNet}, and the reported metric RMSE represent the average of \textbf{five runs}. }
\label{tab:rq1_rmse}
    \centering
    \footnotesize
    \renewcommand{\arraystretch}{1.2}
    \setlength{\tabcolsep}{3pt}
    \resizebox{\textwidth}{!}{
    \begin{tabular}{cc|cccc|cccc}
    \toprule
    \multirow{3}{*}{} & Dataset  & \multicolumn{4}{c|}{\textsc{Pems08} (RMSE $\downarrow$)} & \multicolumn{4}{c}{\textsc{UrbanEV} (RMSE $\downarrow$)}  \\
    \midrule
    & Remaining Ratio \% & 10 & 30 & 50 & 70  & 10 & 30 & 50 & 70 \\ \midrule
    \parbox[t]{4mm}{\multirow{13}{*}{\rotatebox[origin=c]{90}{Static}}}
    
    & Hard Random
    & 31.29\blue{16.28\%} & 29.24\blue{8.66\%} & 28.72\blue{6.73\%} & 28.29\blue{5.13\%}
    & 8.24\blue{22.07\%} & 7.73\blue{14.52\%} & 7.35\blue{8.89\%} & 7.24\blue{7.26\%}
    \\
    
    & CD~\citep{agarwal2020contextual}
    & 31.31\blue{16.35\%} & 29.04\blue{7.92\%} & 28.61\blue{6.32\%} & 28.10\blue{4.42\%}
    & 8.42\blue{24.74\%} & 7.69\blue{13.93\%} & 7.37\blue{9.19\%} & 7.28\blue{7.85\%}
    \\
    
    & Herding~\citep{welling2009herding}
    & 31.15\blue{15.76\%} & 29.16\blue{8.36\%} & 28.48\blue{5.83\%} & 28.04\blue{4.20\%}
    & 8.60\blue{27.41\%} & 7.60\blue{12.59\%} & 7.52\blue{11.41\%} & 7.14\blue{5.78\%}
    \\
    
    & K-Means~\citep{sener2018active}
    & 31.50\blue{17.06\%} & 29.23\blue{8.62\%} & 28.74\blue{6.80\%} & 28.05\blue{4.24\%}
    & 8.41\blue{24.59\%} & 7.78\blue{15.26\%} & 7.38\blue{9.33\%} & 7.21\blue{6.81\%}
    \\
    
    & Least Confidence~\citep{coleman2019selection}
    & 32.36\blue{20.25\%} & 30.11\blue{11.89\%} & 28.54\blue{6.06\%} & 28.43\blue{5.65\%}
    & 8.29\blue{22.81\%} & 8.65\blue{28.15\%} & 7.36\blue{9.04\%} & 7.07\blue{4.74\%}
    \\
    
    & Entropy~\citep{coleman2019selection}
    & 32.46\blue{20.62\%} & 29.01\blue{7.80\%} & 28.48\blue{5.83\%} & 28.18\blue{4.72\%}
    & 8.39\blue{24.30\%} & 7.68\blue{13.78\%} & 7.37\blue{9.19\%} & 7.28\blue{7.85\%}
    \\
    
    & Margin~\citep{coleman2019selection}
    & 32.18\blue{19.58\%} & 29.20\blue{8.51\%} & 28.55\blue{6.09\%} & 28.32\blue{5.24\%}
    & 8.20\blue{21.48\%} & 7.83\blue{16.00\%} & 7.56\blue{12.00\%} & 7.10\blue{5.19\%}
    \\
    
    & Forgetting~\citep{toneva2018empirical}
    & 30.91\blue{14.86\%} & 29.66\blue{10.22\%} & 28.66\blue{6.50\%} & 28.08\blue{4.35\%}
    & 8.22\blue{21.78\%} & 7.74\blue{14.67\%} & 7.38\blue{9.33\%} & 7.21\blue{6.81\%}
    \\
    
    & GraNd~\citep{paul2021deep}
    & 31.08\blue{15.50\%} & 29.01\blue{7.80\%} & 28.70\blue{6.65\%} & 28.22\blue{4.87\%}
    & 8.23\blue{21.93\%} & 7.76\blue{14.96\%} & 7.42\blue{9.93\%} & 7.31\blue{8.30\%}
    \\
    
    & Cal~\citep{margatina2021active}
    & 30.57\blue{13.60\%} & 29.34\blue{9.03\%} & 29.20\blue{8.51\%} & 28.48\blue{5.83\%}
    & 8.29\blue{22.81\%} & 7.74\blue{14.67\%} & 7.36\blue{9.04\%} & 7.20\blue{6.67\%}
    \\
    
    & Glister~\citep{killamsetty2021glister}
    & 32.32\blue{20.10\%} & 30.20\blue{12.23\%} & 29.83\blue{10.85\%} & 29.30\blue{8.88\%}
    & 8.23\blue{21.93\%} & 8.06\blue{19.41\%} & 7.60\blue{12.59\%} & 7.34\blue{8.74\%}
    \\
    
    & GraphCut~\citep{iyer2021submodular}
    & 31.35\blue{16.50\%} & 29.19\blue{8.47\%} & 28.37\blue{5.43\%} & 28.11\blue{4.46\%}
    & 8.47\blue{25.48\%} & 7.67\blue{13.63\%} & 7.44\blue{10.22\%} & 7.15\blue{5.93\%}
    \\
    
    & FaLo~\citep{iyer2021submodular}
    & 31.04\blue{15.35\%} & 29.22\blue{8.58\%} & 28.59\blue{6.24\%} & 28.32\blue{5.24\%}
    & 8.40\blue{24.44\%} & 7.63\blue{13.04\%} & 7.46\blue{10.52\%} & 7.26\blue{7.56\%}
    \\
    
    \midrule
    \parbox[t]{4mm}{\multirow{5}{*}{\rotatebox[origin=c]{90}{Dynamic}}}
    & Soft Random
    & 32.71\blue{21.55\%} & 29.44\blue{9.40\%} & 28.82\blue{7.10\%} & 27.86\blue{3.53\%}
    & 8.29\blue{22.81\%} & 7.45\blue{10.37\%} & 7.19\blue{6.52\%} & 7.24\blue{7.26\%} 
    \\

    & $\epsilon$-greedy~\citep{raju2021accelerating}
    & 29.94\blue{11.26\%} & 29.64\blue{10.14\%} & 28.74\blue{6.80\%} & 27.96\blue{3.90\%}
    & 8.18\blue{21.19\%} & 7.47\blue{10.67\%} & 7.51\blue{11.26\%} & 7.33\blue{8.59\%}
    \\

    & UCB~\citep{raju2021accelerating}
    & 29.93\blue{11.22\%} & 29.87\blue{11.00\%} & 28.69\blue{6.61\%} & 28.04\blue{4.20\%}
    & 8.15\blue{20.74\%} & 7.46\blue{10.52\%} & 7.47\blue{10.67\%} & 7.30\blue{8.15\%}
    \\

    & InfoBatch~\citep{qin2023infobatch}
    & 30.16\blue{12.08\%} & 29.48\blue{9.55\%} & 28.78\blue{6.94\%} & 28.48\blue{5.83\%} 
    & 7.54\blue{11.70\%} & 7.46\blue{10.52\%} & 7.02\blue{4.00\%} & 6.95\blue{2.96\%} 
    \\

    & \model (\textcolor{orange}{Our})
    & \colorbox[HTML]{DAE8FC}{\textbf{28.37\blue{\textbf{5.46}}}} 
    & \colorbox[HTML]{DAE8FC}{\textbf{27.89\blue{\textbf{3.64}}}} 
    & \colorbox[HTML]{DAE8FC}{\textbf{27.61\blue{\textbf{2.60}}}} 
    & \colorbox[HTML]{DAE8FC}{\textbf{27.71\blue{\textbf{2.97}}}}

    & \colorbox[HTML]{DAE8FC}{\textbf{6.91\blue{\textbf{2.37}}}} 
    &  \colorbox[HTML]{DAE8FC}{\textbf{6.89\blue{\textbf{2.07}}}} 
    &   \colorbox[HTML]{DAE8FC}{\textbf{6.75\blue{\textbf{0.00}}}} 
    & \colorbox[HTML]{DAE8FC}{\textbf{6.75\red{\textbf{0.00}}}}

    \\

    \midrule
    \multicolumn{2}{c|}{Whole Dataset} & \multicolumn{4}{c|}{26.91$_{\pm0.23}$} & \multicolumn{4}{c}{6.75$_{\pm0.09}$} \\
    \bottomrule
    \end{tabular}}
\end{table*}

%% file: appendix/5_morediscussion.tex
\section{More Discussion}\label{more_discuss}

\subsection{Current Limitation} 

In this paper, we thoroughly investigate methods for accelerating spatio-temporal training via dynamic data pruning. . Based on the analysis of the characteristics of spatiotemporal data, we propose a \model. While we have made some progress in this area, some limitations remain to be considered:

\ding{182} \textit{Additional Computational Overhead.} Our pruning strategy introduces a small but not negligible additional overhead. While this overhead is negligible when training large-scale spatio-temporal datasets or large backbone models (\eg, Transformers or deep GNNs), it can offset the efficiency gains when applying the model to extremely lightweight models (\eg, simple MLPs) or small datasets.

\ding{183} \textit{Assumptions of Static Spatial Topology.} Current structure scoring mechanisms rely on a consistent set of spatial nodes $N$ to compute spatial variance. For dynamic graph scenarios where nodes frequently appear or disappear (\eg, the changing demand for ride-hailing services within an expanding region), the definition of ''spatial heterogeneity`` becomes ambiguous, potentially requiring a redesign of the scoring function.

\subsection{Future Work}

One lesson learned from our experiments is that the definition of ``hardness'' in spatio-temporal sample is not static but evolves during the training process. While our current method relies on handcrafted heuristics (variance and stationarity intensity) to define importance, we believe that an automated, learnable scoring policy could further adapt to the model's changing needs. Thus, we view exploring reinforcement Learning or meta-learning to automatically learn the optimal scoring and reweighting functions as a long-term goal for future work.